\documentclass[onefignum, onetabnum,onealgnum]{siamonline190516}
\usepackage{lipsum}

\usepackage{endnotes}

\usepackage{titlesec}
\usepackage{titletoc}

\titleclass{\task}{straight}[\section]
\newcounter{task}
\renewcommand{\thetask}{\arabic{task}}
\titleformat{\task}[hang]
    {\normalfont\LARGE\bfseries}{Task \thetask:}{1em}{}

\titleformat*{\task}{\color{header1}\bfseries}

\titlecontents{task}
              [3.8em] 
              {}
              {\contentslabel{2.3em}}
              {\hspace*{-2.3em}}
              {\titlerule*[1pc]{.}\contentspage}

\titlespacing*{\section}{0ex}{1ex}{1ex}
\titlespacing*{\subsection}{0ex}{1ex}{1ex}
\titlespacing*{\paragraph}{0ex}{1ex}{1ex}
\titlespacing*{\subparagraph}{0pt}{1ex}{1ex}
\titlespacing*{\task}{0em}{1ex}{1ex}

\usepackage[sort&compress,comma,square,numbers]{natbib}

\usepackage{amsfonts,amsmath,amssymb}
\usepackage{bbm}

\usepackage{multicol}
\usepackage{enumitem}

\usepackage{hhline}

\usepackage{comment}
\usepackage{todonotes}
\RequirePackage{ulem}

\usepackage{booktabs}
\usepackage{titlesec}
\usepackage{fancyhdr}
\usepackage{fullpage}

\RequirePackage{titlesec}
\RequirePackage[font={small},{singlelinecheck=false}]{caption}
\usepackage[T1]{fontenc}
\usepackage[scaled]{helvet}
\usepackage[scaled=1.10, med]{zlmtt}

\usepackage{algorithm}
\usepackage{algpseudocode}

\newcommand{\Linefor}[2]{%
    \State \algorithmicfor\ {#1}\ \algorithmicdo\ {#2} \algorithmicend\ \algorithmicfor%
}

\providecommand{\sct}[1]{{\sc \texttt{#1}}}

\newcommand{\Sporf}{\sct{Sporf}}

\newcommand{\Rf}{{\sc \texttt{RF}}}
\newcommand{\Ccf}{{\sc \texttt{CCF}}}
\newcommand{\Rrrf}{{\sc \texttt{RR-RF}}}
\newcommand{\Xgboost}{{\sc \texttt{XGBoost}}}
\newcommand{\Frc}{{\sc \texttt{F-RC}}}

\newcommand{\argmax}{\operatornamewithlimits{argmax}}

\providecommand{\mc}[1]{\mathcal{#1}}

\providecommand{\mt}[1]{\widetilde{#1}}

\newcommand{\Real}{\mathbb{R}}

\newcommand{\II}{\mathbb{I}}           

\usepackage{atbegshi}
\usepackage{array}
\usepackage{longtable}
\usepackage{pdflscape}
\usepackage{color}
\usepackage{multirow}
\usepackage{tablefootnote}
\usepackage[title]{appendix}

\title{Sparse Projection Oblique Randomer Forests}

\newcommand*\samethanks[1][\value{footnote}]{\footnotemark[#1]}

\author{
    Tyler M. Tomita%
    \thanks{Department of Psychological and Brain Sciences,
        Johns Hopkins University (\email{ttomita@jhu.edu})}
    \and
    James Browne%
    \thanks{Department of Computer Science,
        Johns Hopkins University}
    \and
    Cencheng Shen%
    \thanks{Department of Applied Economics and Statistics,
        University of Delaware}
    \and
    Jaewon Chung%
    \thanks{Center for Imaging Science,
        Johns Hopkins University}
    \and
    Jesse L. Patsolic%
    \samethanks[4]
    \and
    Benjamin Falk%
    \samethanks[4]
    \and
    Jason Yim%
    \thanks{DeepMind,
        London, UK}
    \and
    Carey E. Priebe%
    \thanks{Department of Applied Mathematics and Statistics,
        Johns Hopkins University}
    \and
    Randal Burns%
    \samethanks[2]
    \and
    Mauro Maggioni%
    \samethanks[6]
    \and
    Joshua T. Vogelstein%
    \thanks{Corresponding author; 
        Department of Biomedical Engineering,
        Institute for Computational Medicine,
        Kavli~Neuroscience~Discovery Institute,
        Johns Hopkins University (\email{jovo@jhu.edu},
        \url{http://www.jovo.me/}).}
}

\begin{document}

\maketitle



\noindent
\textbf{%
Decision forests, including Random Forests  and Gradient Boosting Trees, have recently demonstrated state-of-the-art performance in a variety of machine learning settings. 
Decision forests are typically  ensembles of \emph{axis-aligned} decision trees; that is, trees that split only along feature dimensions. In contrast, many recent extensions to decision forests  are based on axis-oblique splits. Unfortunately, these extensions forfeit one or more of the favorable properties of decision forests based on axis-aligned splits, such as robustness to many noise dimensions, interpretability, or computational efficiency. We introduce yet another decision forest, called ``Sparse Projection Oblique Randomer Forests'' (\Sporf). SPORF  uses  very sparse random projections,  i.e., linear combinations of a small subset of features. \Sporf~significantly improves accuracy over existing state-of-the-art algorithms on a standard benchmark suite for classification with $>100$ problems of varying dimension, sample size, and number of classes. To illustrate how \Sporf~addresses the limitations of both axis-aligned and existing oblique decision forest methods,  we conduct extensive simulated experiments. \Sporf~typically yields improved performance over existing decision forests, while mitigating computational efficiency and scalability and maintaining interpretability. \Sporf~can easily be incorporated into other ensemble methods such as boosting to obtain potentially similar gains.  
}

\section{Introduction}
Over the last two decades, ensemble methods have risen to prominence as the state-of-the-art for general-purpose machine learning tasks. One of the most popular and consistently strong ensemble methods  is Random Forests (\Rf), which uses decision trees as the base learners \citep{Delgado2014,Caruana2008,caruana2006}. More recently, another tree ensemble method known as gradient boosted decision trees (GBTs)  has seen a spike in popularity, largely due to the release of a fast and scalable cross-platform implementation, \Xgboost~\citep{chen2016}. GBTs have been a key component of many Kaggle competition-winning solutions, and was part of the Netflix Prize winning solution \citep{chen2016}.

\Rf~and \Xgboost~are ensembles of ``axis-aligned'' decision trees. With such decision trees, the feature space is recursively split along directions parallel to the coordinate axes. Thus, when classes seem inseparable along any single dimension, axis-aligned splits require very deep trees with complicated step-like decision boundaries, leading to increased variance and over-fitting. 
To address this, Breiman also proposed and characterized Forest-RC (\Frc), which splits on linear combinations of coordinates rather than individual coordinates~\cite{Breiman2001}. 
These so-called ``oblique'' ensembles include the axis-aligned ensembles as a special case, and therefore have an increased expressive capacity, conferring  potentially  better learning properties.  Perhaps because of this appeal, numerous other oblique decision forest methods have been proposed, including the Random Rotation Random Forest  (\Rrrf)~\citep{Blaser2016}, and the Canonical Correlation Forest  (\Ccf)~\cite{Rainforth2015}. 
Unfortunately, these methods forfeit many of the desirable properties that axis-aligned trees possesses, such as computational efficiency, ease of tuning, insensitivity to a large proportion of irrelevant (noise) inputs, and interpretability. Furthermore, while these methods perform much better than axis-aligned ensembles on some problems, they perform much worse than axis-aligned ensembles on some problems for which axis-aligned splits would in fact be highly informative. Therefore, there is a need for a method that combines the expressive capacity of oblique ensembles with the benefits of axis-aligned ensembles.

We propose Sparse Projection Oblique Randomer Forests (\Sporf)  for learning an ensemble of oblique, interpretable, and computationally efficient decision trees. At each node of each tree, \Sporf\ searches for splits over a sample of very sparse random projections~\cite{Li2006}, rather than axis-aligned splits. 
Very sparse random projections preserve many of the desirable properties of axis-aligned decision trees, while mitigating their issues.

In section \ref{section: sporf}, we delineate a set of desirable properties of a decision forest algorithm, and describe how current axis-aligned and oblique decision forest algorithms each fail to possess at least one of these. This motivates a flavor of Sparse Random Projections for randomly sampling candidate split directions. In Section \ref{section: simulations}, we show on simulated data settings how our method possesses all of these desirable properties, while other methods do not. In Section \ref{section: real data} we find that, in practice, our method tends to be more accurate than RF and existing methods on many real data sets. Last, in Section \ref{section:compute} we demonstrate how are method is computationally expedient and scalable.

Our  statistically- and computationally-efficient parallelized  implementations are available from \url{https://neurodata.io/sporf/} in both R and Python.  Our R package is available on the Comprehensive R Archive Network (CRAN) (\url{https://cran.r-project.org/web/packages/rerf/}), and our Python package is available from PyPi (\url{https://pypi.org/project/rerf/2.0.2/}), and is sklearn API compliant. 

\section{Background \& Related Work}
\label{section: background}



%

\subsection{Random Forests}

The original random forest (\Rf) procedure popularized by Leo Breiman is one of the most commonly employed classification learning algorithms~\cite{Breiman2001}. \Rf~proceeds by building $T$ decision trees via a series of recursive binary splits of the training data. The nodes in a tree are split into two daughter nodes by maximizing some notion of information gain, which typically reflects the reduction in class impurity of the resulting partitions. A common measure of information gain in decision trees is the decrease in Gini impurity, $I(S)$, for a set of observations $S$. The Gini impurity for classification is defined as
$I(\mathcal{S}) = \sum_{k = 1}^Kf_k(1 - f_k)$,
where $f_k = \frac{1}{|\mc{S}|}\sum_{i \in \mc{S}} \II[y_i = c_k]$.

More concretely, let $\theta = (j,\tau)$, where $j$ is an index selecting a dimension and $\tau$ is a splitting threshold. Furthermore, let $\mc{S}^L_{\theta} = \{i: x_i^{(j)} \leq \tau, \forall i \in \mc{S}\}$ and $\mc{S}^R_{\theta} = \{i: x_i^{(j)} > \tau, \forall i \in \mc{S}\}$ be the subsets of $\mathcal{S}$ to the left and right of the splitting threshold, respectively.  Here, $x_i^{(j)}$ denotes the value of the $jth$ feature for the $ith$ observation. Let $n_S$, $n_L$, and $n_R$ denote the number of points in the parent, left, and right child nodes, repsectively. A split is made on a "best" $\theta^* = (j^*,\tau^*)$ via the following optimization:
\begin{align*} \label{eq:node}
  \textstyle \theta^* = \argmax\limits_{\theta} n_S I(\mathcal{S}) - n_L I(\mathcal{S}^L_{\theta}) - n_R I(\mathcal{S}^R_{\theta}).
\end{align*}
This optimization is carried out by exhaustively searching for the best split threshold $\tau^*$ over a random subset of the features. Specifically, a random subset of the $p$ features is sampled. For each feature in this subset, the observations are sorted from least to greatest, and the split objective function is evaluated at each midway point between adjacent pairs of observations.

Nodes are recursively split until a stopping criteria is reached. Commonly, the recursion stops when either a maximum tree depth is reached, a minimum number of observations in a node is reached, or a node is completely pure with respect to class label. The result of the tree induction algorithm is a set of split nodes and leaf nodes. The leaf nodes are disjoint partitions of the feature space $\mc{X}$, and each one is associated with a local prediction function. Let $l_m$ be the $m^{th}$ leaf node of an arbitrary classification tree, and let $\mc{S}(l_m) = \{i: x_i \in l_m \forall i \in [n]\}$ be the subset of the training data contained in $l_m$. The local leaf prediction is: 
\begin{align*}
{h(l_m) = \argmax\limits_{c_k \in \mc{Y}}\sum\limits_{i \in \mc{S}(l_m)}\II[y_i = c_k]}
\end{align*}
A tree makes a prediction for a new observation $x$ by passing the observation down the tree according to the split functions associated with each split node until a terminal leaf node is reached. Letting $m(x)$ be the index of the leaf node that $x$ falls into, the tree prediction is $h(l_{m(x)})$. Let $\widehat{y}^{(t)}$ be the prediction made by the $t^{th}$ tree. Then the prediction of the \Rf~is the plurality vote
of the predictions made by each tree:
\begin{align*}
{\widehat{y} = \argmax\limits_{c_k \in \mc{Y}}\sum\limits_{t = 1}^T\II[\widehat{y}^{(t)} = c_k]}
\end{align*}
\noindent 
\citet{Breiman2001} proved that the misclassification rate of a tree ensemble is bounded above by a function inversely proportional to the strength and diversity of its trees.
\Rf~decorrelates (diversifies) the trees via two mechanisms: (1) constructing each tree on a random bootstrap sample of the original training data, and (2) restricting the optimization of the splitting dimension $j$ over a random subset of the total $p$ dimensions. The combination of these two randomizing effects typically leads to generalization performance that is much better than that of any individual tree \citep{Breiman2001}. 

\subsection{Oblique Extensions to Random Forest}

Various tactics have been employed to further promote the strength and diversity of trees. One feature of \Rf~that limits both strength and diversity is that splits must be along the coordinate axes of the feature space. Therefore, one main focus for improving  \Rf~is to somehow relax this restriction. The resulting forests are sometimes referred to as ``oblique'' decision forests, since the splits can be along directions oblique to the coordinate axes. This type of tree was originally developed for computer graphics applications, and is also known as binary space partitioning (BSP) trees. Statistical consistency of BSP trees has been proven for a simplified data-agnostic BSP tree procedure \cite{DevroyeBook}. Various approaches have been proposed for constructing oblique forests. \citet{Breiman2001} proposed the Forest-RC (\Frc) algorithm, which constructs $d$ univariate projections, each projection a linear combination of $L$ randomly chosen dimensions. The weights of each projection are independently sampled uniformly over the interval $[-1,1]$. Strangely, Breiman's \Frc~never garnered the popularity that \Rf~has acquired;  both \citet{Breiman2001} and \citet{Tomita2017} indicate that \Frc~tends to empirically outperform \Rf~on a wide variety of datasets. \citet{Heath1993}  sample a randomly oriented hyperplane at each split node, then iteratively perturb the orientation of the hyperplane to achieve a good split. \citet{Rodriguez2006} attempted to find discriminative split directions via PCA. \citet{Menze2011} perform supervised learning of linear discriminative models at each node. \citet{Blaser2016} proposed the Random Rotation Random Forest (\Rrrf) method, which uniformly randomly rotates the data prior to inducing each tree. Trees are then learned via the typical axis-aligned procedure on the rotated data. \citet{Rainforth2015}'s Canonical Correlation Forests (\Ccf) employ canonical correlation analysis at each split node in order to directly compute split directions that maximally correlate with the class labels. \citet{lee2015}'s Random Projection Forests (RPFs) generates a discriminative image filter bank for head-pose estimation at each split node and compresses the responses using random projections. The key thing to note is that all of these aforementioned oblique methods utilize some flavor of random projections, which we briefly introduce next.


\subsection{Random Projections}

Given a data matrix $\mathbf{X} \in \Real^{n \times p}$, one can construct a random projection matrix $\mathbf{A} \in \Real^{p \times d}$ and multiply it by $\mathbf{X}$ to obtain:
\begin{align*}
\widetilde{\mathbf{X}} = \mathbf{XA} \in \Real^{n \times d}, \quad d \ll \min(n,p).
\end{align*}
If the random matrix entries $a_{ij}$ are i.i.d. with zero mean and constant variance, then the much smaller matrix $\widetilde{\mathbf{X}}$ preserves all pairwise distances of $\mathbf{X}$ with small distortion and high probability\footnote{In classification, preservation of pairwise interpoint distances is not important. Rather, minimizing within-class distances while maximizing between-class distances is what is important. However, we introduce the topic because of its relevance and use in many decision tree algorithms, as is discussed in Section \ref{section: sporf}.}.

Due to theoretical guarantees, random projections are commonly employed as a dimensionality reduction tool in machine learning applications \citep{bingham2001,fern2003,fradkin2003,achlioptas2003,hegde2008}. 
Different probability distributions over the entries lead to different average errors and error tail bounds. \citet{Li2006} demonstrates that \textbf{very sparse random projections}, in which a large fraction of entries in $\mathbf{A}$ are zero, can maintain high accuracy and significantly speed up the matrix multiplication by a factor of $\sqrt{p}$ or more. Specifically, a very sparse random projection matrix is constructed by sampling entries $a_{ij}$ with the following probability distribution:

\[
a_{ij} = 
\begin{cases}
+1 & \text{with prob. } \frac{1}{2s} \\
0 & \text{with prob. } 1 - \frac{1}{s}, \qquad \text{typically }s \gg 3\\
-1 & \text{with prob. } \frac{1}{2s} \\
\end{cases}
\]

\citet{Dasgupta2008-rq} proposed Random Projection Trees, in which they sampled dense random projections in an unsupervised fashion to approximate low dimensional manifolds, and later for  vector quantization~\citep{Dasgupta2009-fa} and nearest neighbor search~\citep{Dasgupta2013-sc}. Our work is inspired by this work, but in a supervised setting. 

\subsection{Gradient Boosted Trees}
Gradient boosted trees (GBTs) are another tree ensemble method commonly used for regression and classification tasks. Unlike in \Rf, GBTs are learned in an iterative stage-wise manner by directly minimizing a cost function via gradient descent \citep{breiman1998,friedman2001}. Despite the obvious differences in the learning procedures between GBT and \Rf, they tend to perform comparably. A study by \citet{Wyner2017-ax} argues that \Rf~ and GBT are both successful for the same reason---namely both are weighted ensembles of interpolating classifiers that learn local decision rules.

GBTs have recently seen a marked surge in popularity, and were used as components in many recent Kaggle competitions. This is in part due to their tendency to be accurate over a wide range of settings. Their popularity and success can also be attributed to the recent release of \Xgboost~\citep{chen2016} and {\sc \texttt{LightGBM}} \citep{ke2017}, both extremely fast and scalable open-source software implementations. Due to the success of GBTs in many data science applications, we compare the \Xgboost~implementation to our methods.


\section{Methods}
\subsection{Sparse Projection Oblique Randomer Forests}
\label{section: sporf}

Extensions of \Rf~are often focused on changing the procedure for finding suitable splits, such as employing a supervised linear procedure or searching over a set of randomly oriented hyperplanes. Such extensions, along with \Rf, simply differ from each other by defining different random projection distributions from which candidate split directions are sampled. Thus, they are different special cases of a general random projection forest.

Specifically, let $\mathbf{X} \in \Real^{n \times p}$ be the observed feature matrix of $n$ samples at a split node, each $p$-dimensional.  Randomly sample a matrix $\mathbf{A} \in \Real^{p \times d}$ from distribution $f_\mathbf{A}$, possibly in a data-dependent or supervised fashion. This matrix is used to randomly project the feature matrix, yielding $\mt{\mathbf{X}} = \mathbf{X} \mathbf{A} \in \Real^{n \times d}$, where $d$ is the dimensionality of the projected space. The search for the best split is then performed over the dimensions in the projected space.
As an example, in \Rf, $\mathbf{A}$ is a random matrix in which each of the $d$ columns has only one nonzero entry, and every column is required to be unique. Searching for the best split over each dimension in this projected subspace amounts to searching over a random subset of the original features.

While the best specification of a distribution over random projections (if one exists) is dataset-dependent, it is unreasonable and/or undesirable to try more than a handful of different cases. Therefore, for general purpose classification we advocate for a default projection distribution based on the following desiderata:

\begin{enumerate}[label=\textbf{\textbullet},wide,labelwidth=!,labelindent=0pt,leftmargin=*]

\item \textbf{Random Search for Splits.} The use of guided (supervised) linear search procedures for computing split directions, such as linear discriminant analysis (LDA), canonical correlation analysis (CCA), or logistic regression (LR), can result in failure to learn good split directions on certain classification problems (for example, the XOR problem). On the other hand, searching over a random set of split directions \emph{can} identify good splits in many of such cases. Furthermore, supervised procedures run the risk of being overly greedy and reduce tree diversity, causing the model to overfit noise (we demonstrate this in high-dimensional settings in Section \ref{section: noise}). This is akin to why in \Rf~it is typically better to evaluate a random subset of features to split on, rather than exhaustively evaluate all features. Lastly, supervised procedures become costly if performed at every split node.
  
\item \textbf{Flexible Sparsity.}  {\Rf}s search for splits over fully sparse, or axis-aligned, projections. Thus, it may perform poorly when no single feature is informative. On the other hand, methods that search for splits within a fully dense randomly projected space, such as {\Rrrf}s, perform poorly in high-dimensional settings for which the signal is contained in a small subset of the features. This is because the large space renders the probability of sampling discriminative random projections very small (we refer the reader to pages 49-50 of \citet{Vershynin2019} for relevant theory of random projections). However, inducing an appropriate amount of sparsity in the random projections increases the probability of sampling discriminative projections in such cases. \citet{Tomita2017} demonstrated that \Frc, which allows control over the sparsity of random projections, empirically performed much better than both \Rf~ and \Rrrf
  
\item \textbf{Ease of Tuning.} {\Rf}s tend to work fairly well out-of-the-box, due to their relative insensitivity to hyperparameter settings \cite{Probst2019}. Unfortunately, existing oblique forests introduce additional hyperparameters to which they are sensitive to. 
  
\item \textbf{Data Insight.} Often times the goal is not simply to produce accurate predictions, but to gain insight into a process or phenomenon being studied. While \Rf~models can have complicated decision rules, Gini importance \citep{Breiman2002} has been proposed as a computationally efficient way to assess the relative contribution (importance) of each feature to the learned model. As is explained in Section \ref{section:importance}, existing oblique forests do not lend themselves well to computation of Gini importance.

\item \textbf{Expediency and Scalability.} Existing oblique forest algorithms typically involve expensive computations to identify and select splits, rendering them less space and time efficient than \Rf, and/or lack parallelized implementations.
\end{enumerate}

With these considerations in mind, we propose a new decision tree ensemble method called Sparse Projection Oblique Randomer Forests (\Sporf). Our proposed method searches for splits over sparse random projections \citep{Li2006}. Rather than sampling $d$ non-zero elements of $\mathbf{A}$ and enforcing that each column gets a single non-zero number (without replacement), as \Rf~does, we relax these constraints and sample $\lceil\lambda p d\rceil$ non-zero numbers from $\{-1,+1\}$ with equal probabilities, where $\lambda \in (0, 1]$ is the density (fraction of nonzeros) of $\mathbf{A}$ and $\lceil \cdot \rceil$ is the ceiling function rounding up to the nearest integer.\footnote{While $\lambda$ can range from zero to one, we only try values from $1/p$ up to $5/p$ in our experiments.} These nonzeros are then distributed uniformly at random in $\mathbf{A}$. 
See Algorithms~\ref{alg:rerftrain} and \ref{alg:rerfsplit} for details on how to grow a \Sporf~decision tree.

\Sporf~addresses the all of the desiderata listed above. The use of sparse random projections with control over the sparsity via $\lambda$ addresses the first two. Additionally, $\lambda$ is the only new hyperparameter to tune relative to \Rf. Breiman's \Frc~has an analogous hyperparameter $L$, which fixes the number of variables in every linear combination. However, we show later that \Sporf~is less sensitive to the choice in $\lambda$ than \Frc~is to the choice in $L$. By keeping the random projections sparse with only two discrete weightings of $\pm 1$, Gini importance of projections can be computed in a straightforward fashion. Last, sparse random projections are cheap to compute, which allows us to maintain computational expediency and scalability similar to that of \Rf.

\subsection{Training and Hyperparameter Tuning}
\label{section: tuning}

Unless otherwise stated, model training and tuning for all algorithms except for \Xgboost~and \Ccf~is performed in the following way. Each algorithm uses 500 trees, which  was empirically determined to be sufficient for convergence of out-of-bag error in
for all methods.  The split objective is to maximize the reduction in Gini impurity.
In all methods, classification trees are  fully grown  unpruned (i.e. nodes are split until pure). While fully grown trees often cause a single tree to overfit, averaging over many uncorrelated trees tends to alleviate overfitting. A recent study suggests that \Rf~is relatively insensitive to its hyperparameters compared to other machine learning algorithms. Furthermore, tuning tree depth is shown to be much less critical than tuning  the number of variables sampled at each split node \cite{Probst2019}.
 Two hyperparameters are tuned via minimization of out-of-bag error.
 The first parameter tuned is $d$, the number of candidate split directions evaluated at each split node. Each algorithm is trained for $d = p^{1/4}$, $p^{1/2}$, $p^{3/4}$, and $p$. Additionally, \Sporf~and \Frc~are trained for $d = p^2$. For \Rf, $d$ is restricted to be no greater than $p$ by definition. The second hyperparameter tuned is $\lambda$, the average sparsity of univariate projections sampled at each split node. The values optimized over for \Sporf~and \Frc~are $\{1/p,\ldots,5/p\}$. Note, for \Rf~$\lambda$ is fixed to $1/p$ by definition, since the univariate projections are constrained to be along one of the coordinate axes of the data.

For \Ccf, the number of trees is 500, trees are fully grown, and the split objective is to maximize the reduction in class entropy (this is the default objective found to perform best by the authors). The only hyperparameter tuned is the number of features subsampled prior to performing CCA. We optimize this hyperparameter over the set $\{p^{1/4}, p^{1/2}, p^{3/4}, p\}$. \Ccf~uses a different observation subsampling procedure called \emph{projection boostrapping} instead of the standard bootstrap procedure. Briefly, in projection bootstrapping, all trees are trained on the full set of training observations. Bootstrapping is instead performed at the node level when computing the canonical correlation projections at each node. Once the projections are computed, the projection and corresponding split threshold that maximizes the reduction in Gini impurity is found using all of the node observations (i.e. not just the bootstrapped node observations). Since there are no out-of-bag samples for each tree, we base the selection of the best value on minimization of a five-fold cross-validation error rate instead.

Hyperparameters in \Xgboost~are tuned via grid search using the R caret package (see Appendix \ref{section: tuning_app} for details).


\section{Simulated Data Empirical Performance}
\label{section: simulations}

In this section we demonstrate, using synthetic classification problems, that \Sporf~addresses the statistical issues listed above. In a sense, \Sporf~bridges the gap between \Rf~and existing oblique methods.

\subsection{\Sporf~and Other Oblique Forests are ``More Consistent'' Than \Rf}\label{section: consistency}

Typically, a proposed oblique forest method is motivated through purely empirical examples. Moreover, the geometric intuition behind the proposed method is rarely clearly provided. Here we take a step towards a more theoretical perspective on the advantage of oblique splits in tree ensembles.

\begin{figure}[h!]
\centering
\centerline{\includegraphics[width=0.8\columnwidth,trim={0in 0in 0 0},clip]{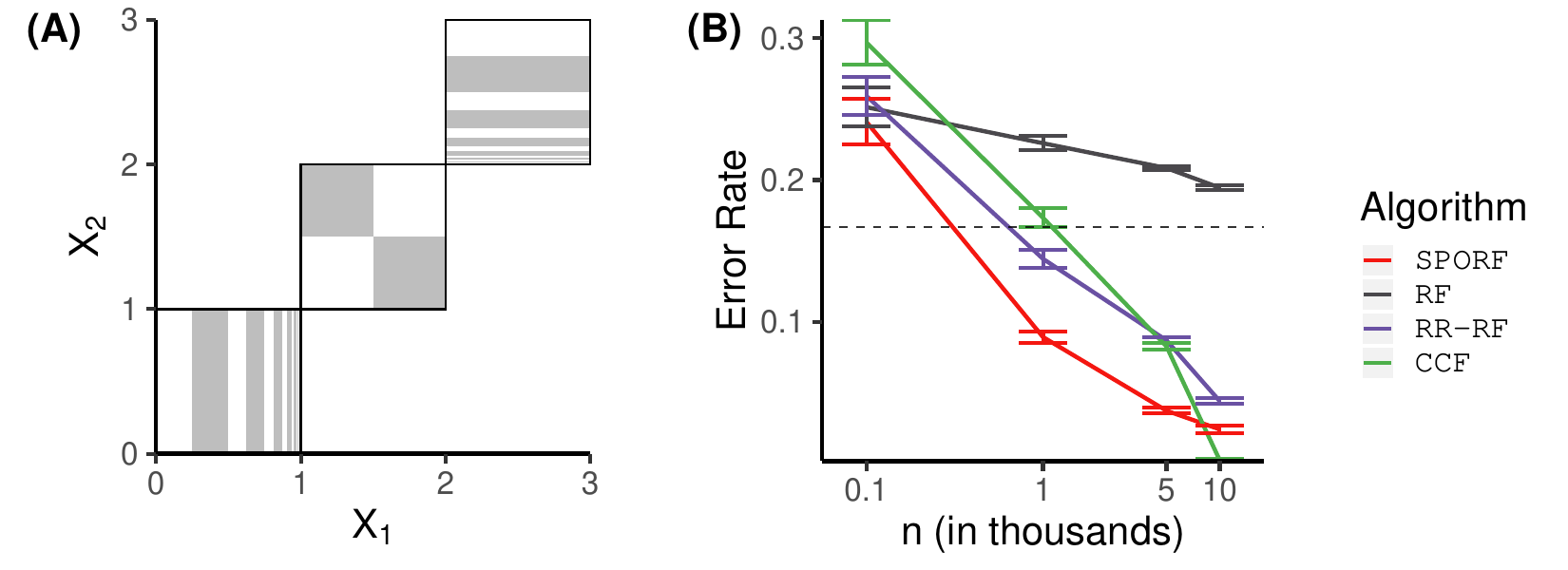}}
\caption{Classification performance on the consistency ($p = 2$) problem as a function of the number of training samples. The consistency problem is designed such that \Rf~has a theoretical lower bound of error of 1/6. \textbf{(A)} The joint distribution of $(X, Y)$. $X$ is uniformly distributed in the three unit squares. The lower left and upper right squares have countably infinite stripes (a finite number of stripes are shown), and the center square is a $2 \times 2$ checkerboard. The white areas represent $f(X) = 0$ and gray areas represent $f(X) = 1$. \textbf{(B)} Error rate as a function of $n$. The dashed line represents the lower bound of error for \Rf, which is 1/6. \Sporf~and other oblique methods achieve an error rate dramatically lower than the lower bound for \Rf.}
\label{fig:consistency_exp}
\end{figure}

Although we do not yet have a proof of the consistency of \Sporf~or other oblique forests, we do propose that they are ''more'' consistent than Breiman's original \Rf. \citet{biau2008} proposed a binary classification problem for which Breiman's \Rf~is inconsistent. The joint distribution of $(X, Y)$ is as follows:
$X \in \Real^2$ has a uniform distribution on $[0,1] \times [0,1] \cup [1,2] \times [1,2] \cup [2,3] \times [2, 3]$. The class label $Y$ is a deterministic function of $X$, that is $f(X) \in \{0, 1\}$. The $[0,1] \times [0, 1]$ square is divided into countably infinite vertical stripes, and $[2, 3] \times [2,3]$ square is similarly divided into countably infinite horizontal stripes. In both squares, the stripes with $f(X) = 0$ and $f(X) = 1$ alternate. The $[1, 2] \times [1, 2]$ square is a $2 \times 2$ checker board. 
Figure \ref{fig:consistency_exp}(A) shows a schematic illustration (because we cannot show countably infinite rows or columns).  On this problem,~\citet{biau2008}  show that \Rf~cannot achieve an error lower than 1/6.  
This is because \Rf~will always choose to split either in the lower left square or top right square and never in the center square. 
On the other hand, Figure \ref{fig:consistency_exp}(B) shows that \Sporf, \Rrrf, and \Ccf~approach perfect classification.  This is due to the fact that, although it is also greedy (i.e. optimizes locally rather than globally), it will choose with some probability oblique splits of the middle square to enable lower error. Therefore, \Sporf~and other oblique methods are empirically more consistent on at least some settings on which \Rf~is neither empirically or theoretically consistent.

To our knowledge, this is the first result comparing the statistical consistency of \Rf~to an oblique forest method. More generally, this result suggests that relaxing the constraint of axis-alignment of splits may allow oblique forests to be  consistent across on a wider set of classification problems. It also highlights why, from a theoretical standpoint, oblique forests are advantageous.

\subsection{Simulated datasets}
\label{section: datasets}

In the sections that follow, we perform a variety of experiments on three carefully constructed simulated classification problems, refered to as \textbf{Sparse Parity}, \textbf{Orthant},  and \textbf{Trunk}. These constructions were chosen to highlight various properties of different algorithms and gain insight into their behavior.

\textbf{Sparse Parity} is a  multivariate generalization of the noisy XOR problem. It is a $p$-dimensional two-class problem in which the class label $Y$ is $0$ if the number of dimensions having positive values amongst the first $p^* < p$ dimensions is even and $Y=1$ otherwise. Thus, only the first $p^*$ dimensions carry information about the class label, and no subsets of dimensions contains any information. Specifically, let $X = (X_1,\ldots,X_p)$ be a $p$-dimensional feature vector, where each $X_1,\ldots,X_p \overset{iid}{\sim} U(-1,1)$. Furthermore, let $Q = \sum_{j=1}^{p^*}{\mathbb{I}(X_j>0)}$, where $p^* < p$ and $\mathbb{I}(X_j > 0)$ is the indicator that the $j^{th}$ feature has a value greater than zero. A sample's class label $Y$ is equal to the parity of $Q$. That is, $Y = \texttt{odd}(Q)$, where $\texttt{odd}$ returns 1 if its argument is odd, and 0 otherwise. The Bayes optimal decision boundary for this problem is a union of hyperplanes aligned along the first $p^*$ dimensions. For the experiments presented in the following sections, $p^* = 3$ and $p = 20$. Figure~\ref{fig:error_synthetic} (A,B) show cross-sections of the first two dimensions taken at two different locations along the third dimension. This setting is designed to be relatively easy for \Frc, but relatively difficult for \Rf.

\textbf{Orthant} is a multi-class problem in which the class label is determined by the orthant in which a datapoint resides. 
An orthant in $\Real^p$ is a generalization of a quadrant in $\Real^2$. In other words, each orthant is a subset of $\Real^p$ defined by constraining each of the $p$ coordinates to be positive or negative. For instance, in $\Real^2$, there are four such subsets: $X = (X_1, X_2)$ can either be in 1) $\Real^+\times \Real^+$, 2) $\Real^- \times \Real^+$, 3) $\Real^- \times \Real^-$, or 4) $\Real^+ \times \Real^-$ . Note that the number of orthants in $p$ dimensions is $2^p$. A key characteristic of this problem is that the individual dimensions are strongly and equally informative. Specifically for our experiments, we sample each $X_1,\ldots,X_p \overset{iid}{\sim} U(-1,1)$. Associate a unique integer index from $1$ to $2^p$ with each orthant, and let $O(X)$ be the index of the orthant in which $X$ resides. The class label is $Y = O(X)$. Thus, there are $2^p$ classes. The Bayes optimal decision boundary in this setting is a union of hyperplanes aligned along each of the $p$ dimensions. We set $p = 6$ in the following experiments. Figure \ref{fig:error_synthetic} (D,E) show cross-sections of the first two dimensions taken at two different locations along the third dimension. This setting is designed to be relatively easy for \Rf~because all optimal splits are axis-aligned. 

\textbf{Trunk} is a balanced, two-class problem in which each class is distributed as a $p$-dimensional multivariate Gaussian with identity covariance matrices \citep{Trunk1979}. Every dimension is informative, but each subsequent dimension is less informative than the last. The class 1 mean is $\mu_1 = (1,\frac{1}{\sqrt{2}},\frac{1}{\sqrt{3}},...,\frac{1}{\sqrt{p}})$, and $\mu_0 = -\mu_1$. The Bayes optimal decision boundary is the hyperplane $(\mu_1 - \mu_0)^\mathsf{T}X = 0$. 
We set $p = 10$ in the following experiments.

\subsection{\Sporf~Combines the Best of Existing Axis-Aligned and Axis-Oblique Methods}
\label{section: simulated performance}

We compare error rates of \Rf, \Sporf, \Frc, and \Ccf~on the sparse parity and orthant problems. Training and tuning are performed as described in Section \ref{section: tuning}. Error rates are estimated by taking a random sample of size $n$, training the classifiers, and computing the fraction misclassified in a test set of 10,000 samples. This is repeated ten times for each value of $n$. The reported error rate is the mean over the ten repeated experiments.

\begin{figure*}[ht!]
\centering
\centerline{\includegraphics[width=\textwidth]{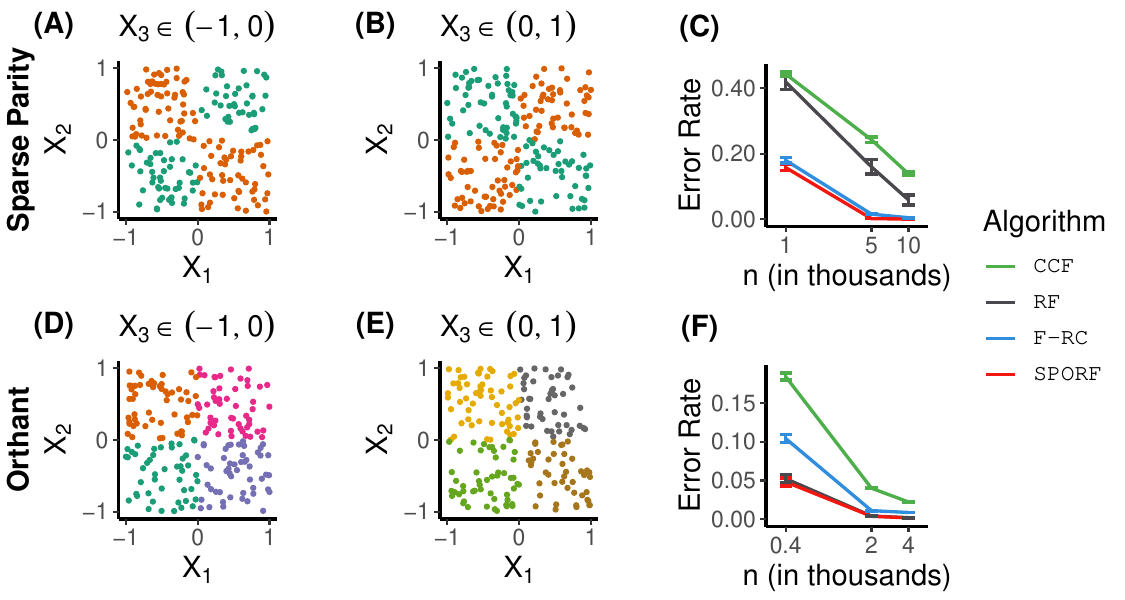}}
\caption{Classification performance on the sparse parity ($p_s = 20$) and orthant ($p_o = 6$) problems for various numbers of training samples. In both case, we sample $X_1, \ldots, X_{p} \overset{iid}{\sim} U(-1, 1)$.
\Frc~has been known to perform much better than \Rf~on the sparse parity problem \citep{Tomita2017}. The orthant problem is designed for \Rf~to perform well because the optimal splits are axis-aligned. \textbf{(A)} A cross-section of the first two dimensions of sparse parity when $X_3 \in (-1, 0)$.  Only the first three dimensions are informative w.r.t. class label. \textbf{(B)} The same as (A), except that the cross-section is taken over $X_3 \in (0, 1)$. \textbf{(C)} Error rate plotted against the number of training samples for sparse parity. Error rate is the average over ten repeated experiments. Error bars indicate the standard error of the mean. \textbf{(D)}-\textbf{(F)} Same as (A)-(C) except for the orthant problem. 
\Sporf~is the only method of the four that performs well across all simulated data settings.}
\label{fig:error_synthetic}
\end{figure*}

\Sporf~performs as well as or better than the other algorithms on both the sparse parity (Figure \ref{fig:error_synthetic}C) and orthant problems (Figure \ref{fig:error_synthetic}F). \Rf~performs relatively poorly on the sparse parity problem. Although the optimal decision boundary is a union of axis-aligned hyperplanes, each dimension is completely uninformative on its own. Since axis-aligned partitions are chosen one at a time in a greedy fashion, the trees in \Rf~struggle to learn the correct partitioning. On the other hand, oblique splits are informative, which substantially improves the generalizability of \Sporf~and \Frc. While \Frc~performs well on the sparse parity problem, it performs much worse than \Rf~and \Sporf~on the orthant problem. On the orthant problem, in which \Rf~is is designed to do exceptionally well, \Sporf~performs just as well. \Ccf~performs poorly on both problems, which may be because CCA is not optimal for the particular data distributions. For instance, in the sparse parity problem, the projection found by CCA at the first node is approximately the difference in class-conditional means, which is zero. Furthermore, \Ccf~only evaluates $d = min(l,C-1)$ projections at each split node, where $l$ is the number of dimensions subsampled and $C$ is the number of classes. On the other hand, \Sporf~evaluates $d$ random projections, and $d$ could be as large as $3^p$ (each of the $p$ elements can be either $0$ or $\pm 1$. Overall, \Sporf~is the only method of the four that performs relatively well on all of the simulated data settings.

\subsection{\Sporf~is Robust to Hyperparameter Selection}
\label{section: robust}

\begin{figure}[ht!]
\centering
\centerline{\includegraphics[width=0.6\columnwidth,trim={0in 0 0 0},clip]{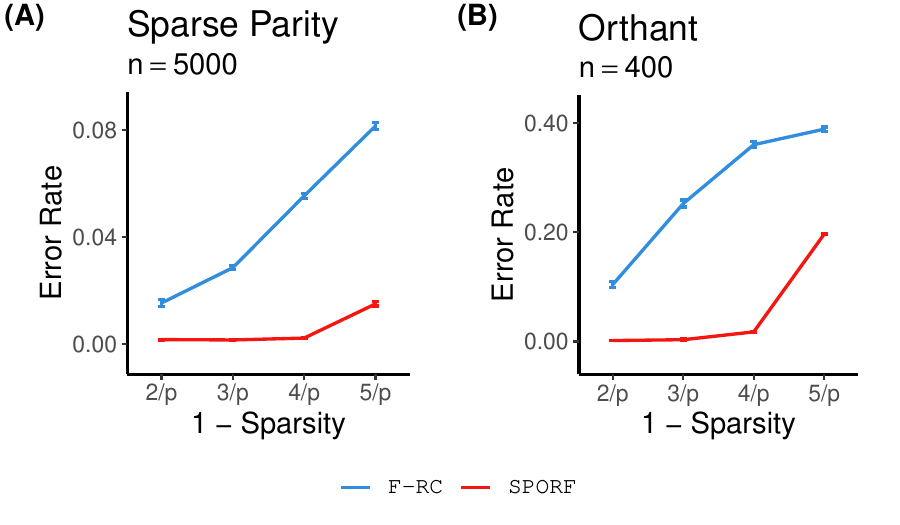}}
\caption{Dependence of error rate on the hyperparameter $\lambda$, which controls the average density (1 - sparsity) of projections for two different simulation settings. \textbf{(A)} Error rate as a function of $\lambda$ on sparse parity ($n = 5000$, $p = 20$). \textbf{(B)} The same as (A) except on orthant ($n = 400$, $p = 6$). In both cases, \Sporf~is less sensitive to different values of $\lambda$ than is \Frc.}
\label{fig:sensitivity_synthetic}
\end{figure}

One key difference between the random projection distribution of \Sporf~and \Frc~is that \Frc~requires that a hyperparameter be specified to fix the sparsity of the sampled univariate projections (i.e., individual linear combinations). Breiman denoted this hyperparameter as $L$. \Sporf~on the other hand, requires that sparsity be specified on the entire random matrix $\mathbf{A}$, and hence, only an \emph{average} sparsity on the univariate projections (details are in Section \ref{section: sporf}). In other words, \Sporf~induces a probability distribution with positive variance on the sparsity of univariate projections, whereas in \Frc~that distribution is a point mass. If the Bayes optimal decision boundary is locally sparse, mis-specification of the hyperparameter controlling the sparsity of $\mathbf{A}$ may be more detrimental to \Frc~than \Sporf. Therefore, we examine the sensitivity of classification performance of \Sporf~and \Frc~to the sparsity hyperparameter $\lambda$ on the simulated datasets described previously. For \Sporf, $\lambda$ is defined as in Section \ref{section: sporf}. For \Frc, we note that $\lambda = L/p$ (i.e. density of a univariate projection is the number of features to combine, divided by the total number of features). For each of $\lambda \in \{\frac{2}{p},\ldots,\frac{5}{p}\}$, the best performance for each algorithm is selected with respect to the hyperparameter based on minimum out-of-bag error. Error rate on the test set is computed for each of the four hyperparameter values for the two algorithms. Figure \ref{fig:sensitivity_synthetic} shows the dependence of error rates of \Sporf~and \Frc~on $\lambda$ for the Sparse Parity ($n$ = 5,000) and Orthant ($n = 400$) settings. The $n$ = 5,000 setting for Sparse Parity was chosen because both \Frc~and \Sporf~perform well above chance (see Figure \ref{fig:error_synthetic}C). The $n$ = 400 setting for Orthant was chosen for the same reason and also because it displays the largest difference in classification performance in Figure \ref{fig:error_synthetic}F. In both settings, \Sporf~is more robust to the choice of sparsity level than \Frc.

\subsection{\Sporf~Learns Important Features}
\label{section:importance}

For many data science applications, understanding which features are important is just as critical as finding an algorithm with excellent predictive performance. One of the reasons \Rf~is so popular is that it can learn good predictive models that simultaneously lend themselves to extracting suitable feature importance measures. One such measure is the mean decrease of Gini importance (hereafter called Gini importance) \citep{DevroyeBook}. This measure of importance is popular because of its computational efficiency: it can be computed during training with minimal additional computation. For a particular feature, it is defined as the sum of the reduction in Gini impurity over all splits of all trees made on that feature. With this measure, features that tend to yield splits with relatively pure nodes will have large importance scores. When using \Rf, features with low marginal information about the class label, but high pairwise or other higher-order joint distributional information, will likely receive relatively low importance scores. Since splits in \Sporf~are linear combinations of the original features, such features have a better chance of being identified.  For \Sporf, we compute Gini importance for each unique univariate projection (i.e. single linear combination). Of note, two projections that differ only by a sign, project into the same  subspace. However, in the experiment that follows we do not check whether any two projections used in the grown forest differ only by a sign.

Another measure of feature importance which we do not consider here is the permutation importance. Permutation importance of a particular feature is computed by shuffling the values along that feature and subsequently assessing how much the error rate increases using the shuffled feature to predict (relative to intact). This measure is considerably slower to compute than is Gini importance in high-dimensional settings because predictions are made for each permuted feature. Furthermore, it is unclear to us how to appropriately compute permutation importance for linear combinations of features.

Another advantage of \Sporf is that methods such as \Frc~,\Rrrf~and \Ccf~do not lend themselves 
to computation of Gini importance. The reason for this is that a particular univariate projection must be sampled and chosen many times over many trees in order to compute a stable estimate of its Gini importance. Since the aforementioned algorithms randomly sample continuous coefficients, it is extremely improbable that the same exact univariate projections will be sampled more than once across trees. 
On the other hand, the projections sampled in \Sporf~are sparse and only contain coefficients of $\pm1$, making it much more likely to sample any given univariate projection repeatedly. Furthermore, \Rrrf~and \Ccf~split on dense univariate projections, which tend to be uninterpretable.

\begin{figure}[ht!]
\centering
\includegraphics[width=0.7\columnwidth]{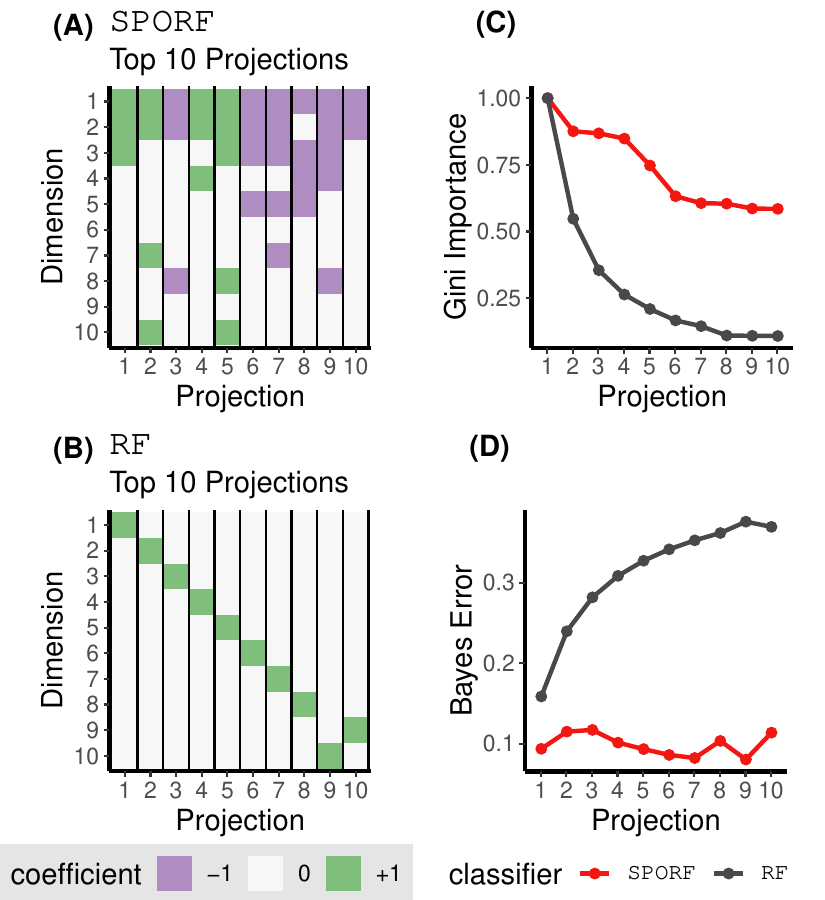}
\caption{The ten projections with the highest Gini importance found by \Rf~and \Sporf~on the Trunk problem with $p = 10, n = 1000$. \textbf{(A)} Visual representation of the top 10 projections identified by \Sporf. The x-axis indicates the projection. The y-axis indicates the index of the ten canonical dimensions. The colors in the heat map indicate the linear coefficients of each canonical dimension that define each of the projections. \textbf{(B)} The same as (A), except for \Rf. \textbf{(C)} Comparison of the Gini importances of the 10 best projections found by each algorithm. \textbf{(D)} Comparison of the Bayes error rate of the 10 best projections found by each algorithm. The top 10 projections used in \Sporf~all have substantially lower Bayes error than those used in \Rf, indicating that \Sporf~learns interpretable informative features.}
\label{fig:importance}
\end{figure}

Gini importance was computed for each feature for both \Rf~and \Sporf~on the Trunk problem with $n = 1,000$. Figure \ref{fig:importance} depicts the features that define each of the top ten split node projections for \Sporf~(A) and \Rf~(B). Projections are sorted from highest  to lowest Gini importance. The top ten projections in \Sporf~are all linear combinations of dimensions, whereas in \Rf~the projections can only be along single dimensions. 
The linear combinations in \Sporf~tend to include the first few dimensions, which contain most of the ``true'' signal. The best possible projection that \Sporf~could sample is the vector of all ones. However, since $\lambda = 1/2$ for this experiment, the probability of sampling such a dense projection  is almost negligible. Figure \ref{fig:importance}(C) shows the normalized Gini importance of the top ten projections for each algorithm. 
The top ten most important features according to \Sporf~are all more important (in terms of Gini) than any of the \Rf~features, except the very first one.  
Figure \ref{fig:importance}(D) shows the Bayes error rate of the top ten projections for each algorithm. 
Again, the top ten features according to \Sporf~are more informative than any of those according to \Rf. In other words, \Sporf~learns features that are more important than any of the observed features, and those features are interpretable, as they are sparse linear combinations of the observed features. The ability of \Sporf~to learn new identifiable features distinguishes it from \Rf, which cannot learn new features.

\section{Real Data Empirical Performance}
\label{section: real data}

\subsection{\Sporf~Exhibits Best Overall Classification Performance on  a Large Suite of Benchmark Datasets}
\label{section: benchmark}

\begin{figure}[ht!]
\centering
{\includegraphics[width=\columnwidth]{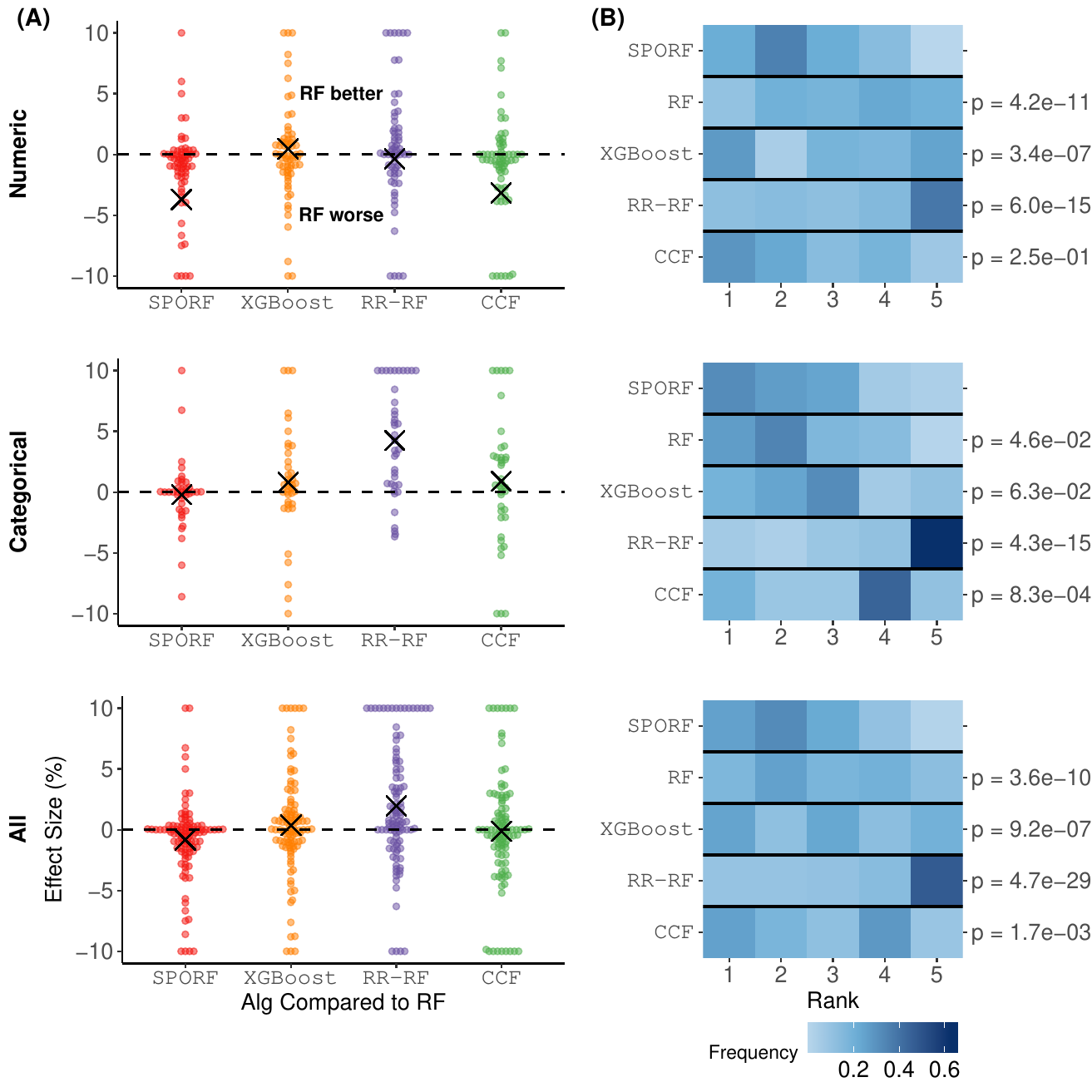}}
\caption{Pairwise comparisons of \Rf~with \Sporf, \Xgboost, \Rrrf, and \Ccf~on the numeric datasets (top), categorical datasets (middle), and all datasets (numeric and categorical combined; bottom) from the UCI Machine Learning Repository (105 datasets total).  \textbf{(A)} Beeswarm plots showing the distributions of classification performance relative to \Rf~for various decision forest algorithms. Classification performance is measured by effect size, which is defined as $\kappa(\Rf) - \kappa(\mathcal{A})$, where $\kappa$ is Cohen's kappa and $\mathcal{A}$ is one of the algorithms compared to \Rf. Each point corresponds to a particular data set. Mean effect sizes are indicated by a black ''x.'' A negative value on the y-axis indicates \Rf~performed worse than a particular algorithm. \textbf{(B)} Histograms of the relative ranks of the different algorithms, where a rank of 1 indicates best relative classification performance and 5 indicates worst. Color indicates frequency, as fraction of data sets. P-values correspond to testing that \Rf, \Xgboost, \Rrrf, and \Ccf~performed worse than \Sporf, using one-sided Wilcoxon signed-rank tests. Overall, \Sporf~tends to perform better than the other algorithms. }\label{fig:error_benchmark}
\end{figure}

\Sporf~compares favorably to \Rf, \Xgboost, \Rrrf, and \Ccf~on a suite of 105 benchmark datasets from the UCI machine learning repository (Figure~\ref{fig:error_benchmark}). 
This benchmark suite is a subset  of the same problem sets previously used to conclude that \Rf~outperformed $>$100 other algorithms~\citep{Delgado2014} (16 were excluded for various reasons such as lack of availability; see Appendix~\ref{sec:benchmarks} for preprocessing details).

Figure \ref{fig:error_benchmark}(A) shows pairwise comparisons of \Rf~with \Sporf~(red), \Xgboost~(yellow), \Rrrf~(purple), and \Ccf~(green) on the UCI datasets. Specifically, let $\kappa(\cdot)$ denote Cohen's kappa (fractional decrease in error rate over the chance error rate) for a particular classification algorithm. Here, error rates are estimated for each algorithm for each dataset via five-fold cross-validation. Error rates for each data set are reported in Appendix \ref{app: tables}. Let $\Delta(\mathcal{A}) = \kappa(\Rf) - \kappa(\mathcal{A})$ be the difference between $\kappa$ for some algorithm $\mc{A}$---either \Sporf, \Xgboost, \Rrrf, or \Ccf---with $\kappa(\Rf)$. Each beeswarm plot in \ref{fig:error_benchmark}(A) represents the distribution of  $\Delta(\mathcal{A})$, denoted "Effect Size," over data sets. Comparisons are shown for the 65 numeric datasets (top), the 40 datasets having at least one categorical feature (middle), and all 105 datasets (bottom). A positive value on the x-axis indicates that \Rf~performed better than the algorithm it is being compared to on a particular dataset, while a negative value indicates it performed worse. Values on the y-axis greater than 10\% were squashed to 10\% and values less than -10\% were squashed to -10\% in order to improve visualization. Mean values are indicated by a black "x." As indicated by the downward skewing distribution, \Sporf~tends to outperform \Rf~over all datasets, due in particular to its relative performance on the numeric datasets. \Rrrf~and \Ccf~also tend to perform similar to or better than \Rf~on the numeric datasets, but unlike~\Sporf~they perform worse than \Rf~on the categorical datasets; the oblique methods are likely sensitive to the one-hot encoding of categorical features. $\kappa$ values for individual data sets and algorithms can be found in Table \ref{tab:class}.

Additionally, we examined how frequently each algorithm ranked in terms of $\kappa$ across the datasets. A rank of one indicates first place (best) on a particular dataset and a rank of five indicates last place (worst). Histograms (in fraction of data sets) of the relative ranks are shown in Figure \ref{fig:error_benchmark}(B). Overall, \Sporf~tends to outperform the other algorithms. This is despite the fact that \Xgboost~is tuned significantly more than \Sporf~in these comparisons (see Section~\ref{section: tuning} for details). Surprisingly, we find that \Rrrf, one of the most recent methods to be proposed, has a strong tendency to perform the worst. One-sided Wilcoxon signed-rank tests were performed to determine whether \Sporf~performed significantly better than each of the other algorithms. Specifically, the null hypothesis was that the median $\kappa$ value of \Sporf~is greater than that of each algorithm being compared to. P-values are shown for each algorithm compared with \Sporf~to the right of each histogram in Figure \ref{fig:error_benchmark}(B). Over all data sets, we found that  p-values were $< 0.005$ for every algorithm compared with \Sporf.
 

\subsection{Identifying Default Hyperparameter Settings}
\label{section: hyperparameters}

\begin{figure}[ht!]
\centering
{\includegraphics[width=0.6\columnwidth]{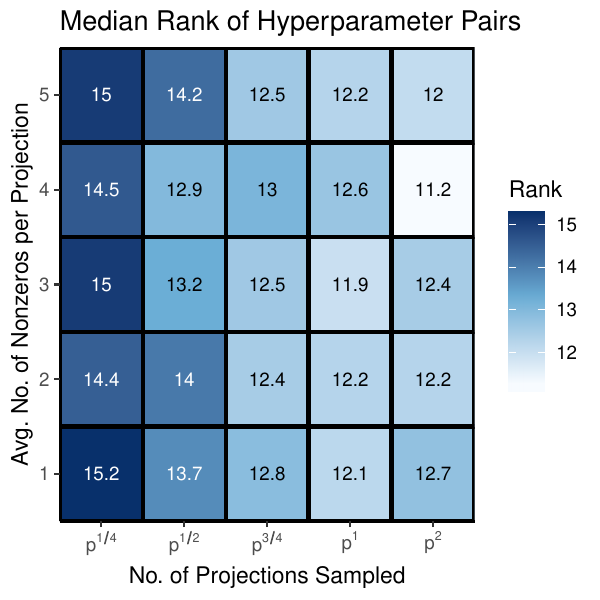}}
\caption{Median rank of \Sporf's ($d$, $\lambda$) hyperparameter pairs on the UCI classification datasets (lower is better).  Although $(p^2, 4/p)$ is the  best performance-wise, we select $(p, 3/p)$ as the default
because of a good balance between accuracy and training time.}\label{fig:hyperparameters}
\end{figure}

While the hyperparameters $\lambda$ and $d$ of \Sporf~were tuned in this comparison, default hyperparameters can be of great value to researchers who use \Sporf~out of the box. This is especially true for those not familiar with the details of a particular algorithm or those having limited time and computational budget. Therefore, we sought suitable default values for $\lambda$ and $d$ based on classification performance on the UCI datasets. For each dataset, for each fold the hyperparameter settings are ranked based on Cohen's kappa computed on the held out set. A rank of $n$ indicates $n^{th}$ place (i.e. first place indicates largest kappa). Ties in the ranking procedure are handled by assigning all ties the same averaged rank. For example, consider the set of real numbers $\{a_1, a_2, a_3\}$ such that $a_1 > a_2 = a_3$. Then $a_1$ would be assigned a rank of three and $a_2$ and $a_3$ would both be assigned a rank of $(1 + 2)/2 = 1.5$. The rank of each hyperparameter pair was averaged over the five folds. Finally, for each hyperparameter pair, the median rank is computed over the datasets. The median rank for each hyperparameter setting is depicted in Figure \ref{fig:hyperparameters}. The results here suggest that $d = p^2$ and $\lambda = 4/p$ is the best default setting for \Sporf~with respect to classification performance. However, we choose the setting $d = p$ and $\lambda = 3/p$ as the default values in our implementation because it requires substantially less training time for moderate to large $p$ at the expense of only a slightly greater tendency to perform worse on the UCI datasets.

\subsection{\Sporf~is Robust to High-Dimensional Noise}
\label{section: noise}

Next, we investigated the effect of adding a varying number of noise dimensions to the UCI benchmark datasets. For each of the 105 UCI datasets used in the previous experiment, $D_{noise}$ standard Gaussian dimensions were appended to the input matrix, for $D_{noise} \in \{10, 100, 1000\}$. Algorithm comparisons were then performed in the same way as before.

\begin{figure*}[ht!]
\centering
\centerline{\includegraphics[width=0.7\textwidth]{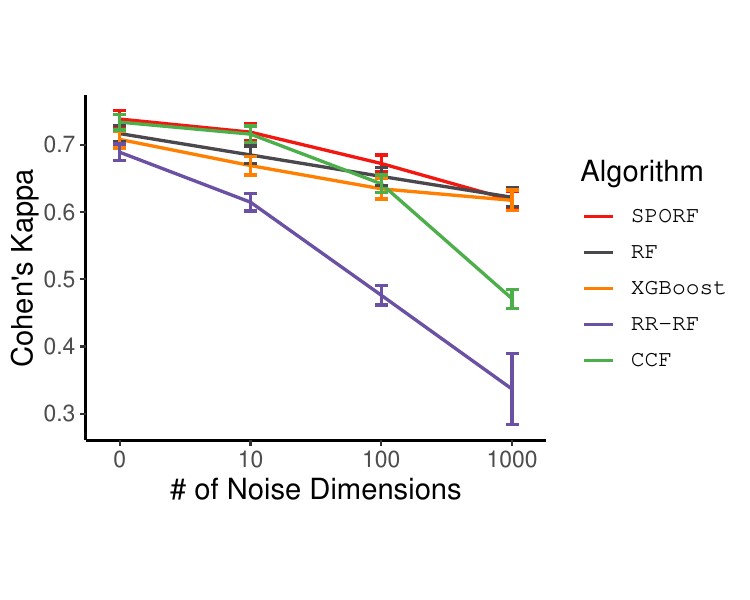}}
\caption{Comparison of classification performance on the UCI benchmark datasets with a varying number of Gaussian noise dimensions added. The x-axis represents the number of noise dimensions added. The y-axis represents the average of Cohen's kappa value over all datasets ($\pm SEM$). \Sporf~is always tied for the best performance. \Ccf~and \Rrrf~are more sensitive to additional noise dimensions.}
\label{fig:benchmark_noise}
\end{figure*}

Figure \ref{fig:benchmark_noise} shows the overall classification performance of \Sporf, \Rf, \Xgboost, and \Ccf~for each value of $D_{noise}$. Each of the points plotted represents the mean Cohen's kappa ($\pm$ SEM) over all datasets. For all values of $D_{noise}$, \Sporf~ties for best classification performance. Notably, \Ccf~performs about as well as \Sporf~when there is little or no additional noise, but degrades substantially when many noise dimensions are added. This suggests that using supervised linear procedures to compute splits may lead to poor out-of-sample performance, likely because the learned features have overfit to the noise dimensions. \Rrrf~degrades even more rapidly than does \Ccf~with increasing numbers of noise dimensions. This can be explained by the fact that features derived from random rotations, which are \emph{dense} linear projections, have very low probability of being informative in the presence of many noise dimensions.

\section{Computational Efficiency and Scalability of \Sporf}
\label{section:compute}

Computational efficiency and scalability are often as  important as accuracy in the choice of machine learning algorithms, especially for big data. In this section we demonstrate that \Sporf, with an appropriate choice of hyperparameter settings, scales similarly to \Rf~with respect to sample size and number of features. Furthermore, we show that our open source implementation is computationally competitive with leading implementations of decision tree ensemble algorithms.

\subsection{Theoretical Time Complexity}
\label{section: time}

The time complexity of an algorithm characterizes how the theoretical processing time for a given input relies on both the hyper-parameters of the algorithm and the characteristics of the input. Let $T$ be the number of trees, $n$ the number of training samples, $p$ the number of features in the training data, and $d$ the number of features sampled at each split node. The average case time complexity of constructing an \Rf~is $\mathcal{O}(Tdn\log^2{n})$ \citep{Louppe2014}. The $dn\log{n}$ accounts for the sorting of $d$ features at each node. The additional $\log{n}$ accounts for both the reduction in node size at lower levels of the tree and the average number of nodes produced. \Rf's near linear complexity shows that a good implementation will scale nicely with large input sizes, making it a suitable algorithm to process big data. \Sporf's average case time complexity is similar to \Rf's, the only difference being that there is an additional term representing a sparse matrix multiplication that is required in each node. This makes \Sporf's complexity $\mathcal{O}(Td n \log^2{n}+ Td n p \lambda))$, where $\lambda$ is the fraction of nonzeros in the $p\times d$ random projection matrix. We generally let $\lambda$ be close to $1/p$, giving a complexity of $\mathcal{O}(Tdn\log^2{n})$, which is the same as for \Rf. Of note, in \Rf~$d$ is constrained to be no greater than $p$, the dimensionality of the data. \Sporf, on the other hand, does not have this restriction on $d$. Therefore, if $d$ is selected to be greater than $p$, \Sporf~may take longer to train. However, $d > p$ often results in improved classification performance.

\subsection{Theoretical Space Complexity}

The space complexity of an algorithm describes how the theoretical maximum memory usage during runtime scales with the number of inputs and hyperparameters. Let $c$ be the number of classes and $T$, $p$, and $n$ be defined as in Section \ref{section: time}. 
Building a single tree requires the data matrix to be kept in memory, which is $\mathcal{O}(np)$. During an attempt to split a node, two $c$-length arrays store the counts of each class to the left and to the right of the candidate split point. These arrays are used to evaluate the decrease in Gini impurity or entropy. Additionally, a series of random sparse projection vectors are sequentially assessed. Each vector has less than $p$ nonzeros. Therefore this term is dominated by the $np$ term. Assuming trees are fully grown, meaning each leaf node contains a single data point, the tree has $2n$ nodes in total. This term gets dominated by the $np$ term as well. Therefore, the space complexity to build a \Sporf~is $\mathcal{O}(T(np + c))$. This is the same as that of \Rf.

%
%

\subsection{Theoretical Storage Complexity}
Storage complexity is the  disk space required to store a forest, given the inputs and hyperparameters. Assume that trees are fully grown. For each leaf node, only the class label of the training data point contained within the node is stored, which is $\mathcal{O}(1)$. For each split node, the split dimension index and threshold are stored, which are also both $\mathcal{O}(1)$. Therefore, the storage complexity of a \Rf~is $\mathcal{O}(Tn)$.

For a \Sporf, the only aspect that differs is that a (sparse) vector is stored at each split node rather than a single split dimension index. 
Let $z$ denote the average number of nonzero entries in a vector projection stored at each split node. Storage of this vector at each split node requires $\mathcal{O}(z)$ memory. Therefore, the storage complexity of a \Sporf~is $\mathcal{O}(Tnz)$.  $z$ is a random variable whose prior is governed by $\lambda$, which is typically set to $1/p$.  The posterior mean of $z$ is determined also by the data; empirically it is close to $z = 1$. Therefore, in practice, the storage complexity of \Sporf~is close to that of \Rf. 

\subsection{Empirical Computational Efficiency and Scalability}

\subsection{Implementation Details}
\label{section: implementation}
We use our own R implementation for evaluations of \Rf, \Sporf, \Frc, and \Rrrf~\citep{rerf}. It was more difficult to modify one of the existing popular tree learning implementations due to the particular way in which they operate on the input data. In all of the popular axis-aligned tree learning implementations, each feature in the input data matrix is sorted just once prior to inducing a tree, and the tree induction procedure operates directly on this presorted data. Since trees in a \Sporf~include  splitting on new features consisting of linear combinations of the original features, pre-sorting the data is not an option. Therefore our implementation is written from scratch in mostly native R. The code has been extensively profiled and optimized for speed and memory performance. Profiling revealed the primary performance bottleneck to be the portion of code responsible for finding the best split. In order to improve speed, this portion of code was implemented in C++ and integrated into R using the Rcpp package \citep{Rcpp}. Further speedup is achieved through multicore parallelization of tree construction and byte-compilation via the R compiler package.

\Xgboost~is evaluated using the R implementation available on CRAN \citep{xgboost}. \Ccf~is evaluated using the authors' openly available MATLAB implementation \citep{Rainforth2015}.

\subsubsection{Comparison of Algorithms Using the Same Implementation}

\begin{figure}[ht!]
\centering
{\fontfamily{lmss}\selectfont 
\textbf{Sparse Parity}
}
\centerline{\includegraphics[width=0.8\columnwidth]{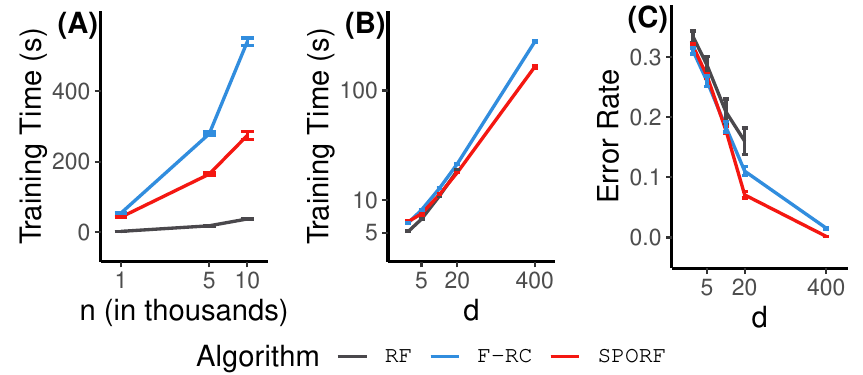}}
\caption{Comparison of training times of \Rf, \Sporf, and \Frc~on the 20-dimensional sparse parity setting. \textbf{(A)} Dependency of training time using the best set of hyperparameters  (y-axis) on the number of training samples (x-axis) for the sparse parity problem. \textbf{(B)} Dependency of training time (y-axis) on the number of projections sampled at each split node (x-axis) for the sparse parity problem with $n =5000$. \textbf{(C)} Dependency of error rate (y-axis) on the number of projections sampled at each split node (x-axis) for the sparse parity problem with $n=5000$. \Sporf~and \Frc~can sample many more than $p$ projections, unlike \Rf. As seen in panels (B) and (C), increasing $d$ above $p$ meaningfully improves classification performance at the expense of larger training times. However, comparing error rates and training times at $d = 20$, \Sporf~can classify substantially better than \Rf~even with no additional cost in training time.}
\label{fig:tradeoff}
\end{figure}

Figure \ref{fig:tradeoff}(A)
shows the training times of \Rf, \Frc, and \Sporf~on the sparse parity problem. The reported training times  correspond to the best hyperparameter settings for each algorithm. Experiments are run  using an Intel Xeon E5-2650 v3 processors clocked at 2.30GHz with 10 physical cores, 20 threads, and 250 GB DDR4-2133 RAM. The operating system is Ubuntu 16.04. \Frc~is the slowest, \Rf~is the fastest, and \Sporf~is in between. While not shown, we note that a similar trend holds for the orthant problem. Figure~\ref{fig:tradeoff}(B) shows that when the hyperparameter $d$ of \Sporf~and \Frc~is the same as that for \Rf, training times are comparable. However, training time continues to increase as $d$ exceeds $p$ for \Sporf~and \Frc, which largely accounts for the trend seen in Figure~\ref{fig:tradeoff}(A). Figure~\ref{fig:tradeoff}(C) indicates that this additional training time comes with the benefit of substantially improved accuracy. Restricting $d$ to be no greater than $p$ for \Sporf~in this setting  would still perform noticeably better than \Rf~at no additional cost in training time. Therefore, \Sporf~does not trade off accuracy for time. Rather, for a fixed computational budget, it achieves better accuracy, and if allowed to use more computation, further improves accuracy.

\subsubsection{Comparison of Training and Prediction Times for Different Implementations}

\begin{figure}[ht!]
\centering
\includegraphics[width=0.8\columnwidth]{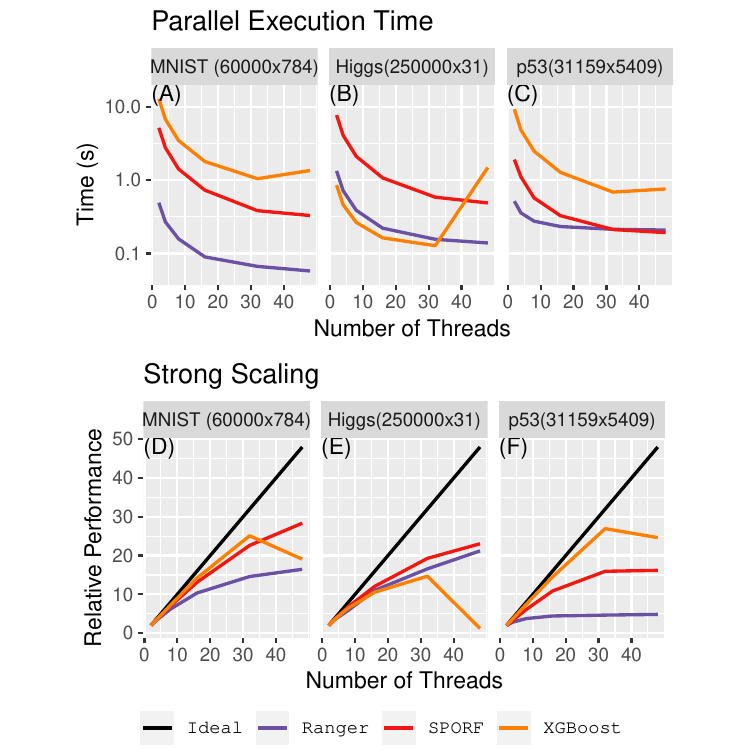}
\caption{
\textbf{(A-C)} The per-tree training time for three large real world datasets. Training was performed using matching parameters where possible, and default parameters otherwise. \Sporf's performance---even though it is written mostly in native R, as compared to the other optimized C++ codes---is comparable to the highly optimized \Xgboost~and Ranger and even outperforms \Xgboost~on two of the datasets.  
\textbf{(D-F):} Strong scaling is the time needed to train a forest with one core divided by the time needed to train a forest with multiple cores. This is a measurement of a system's ability to efficiently utilize additional resources. \Sporf~is able to scale well over the entire range of tested cores, whereas \Xgboost~has sharp drops in scalability during which it is unable to use additional threads due to characteristics of the given datasets.  The p53 dataset, despite having a large number of dimensions, is easily classifiable, which leads to short trees.  The p53 strong scaling plot shows that when trees are short, the overhead of multithreading prevents \Sporf~from efficiently using the additional resources.
}
\label{fig:training}
\end{figure}

We developed and maintain an open  multi-core R implementation of \Sporf~, which is hosted on CRAN \citep{rerf}. We compare both speed of training and strong scaling of our implementation to those of the R Ranger \citep{ranger} and \Xgboost~\citep{xgboost} packages, which are currently two of the fastest, parallelized decision tree ensemble software packages available. Strong scaling is the time needed to train a forest with one core divided by the time needed to train a forest with multiple cores. Ranger offers a fast multicore version of \Rf~that has been extensively optimized for runtime performance. \Xgboost~offers a fast multicore version of gradient boosted trees, and computational performance is optimized for shallow trees. Both Ranger and \Xgboost~are C++ implementations with R wrappers, whereas our \Sporf~implementation is almost entirely native R.
Hyperparameters are chosen for each implementation so as to make the comparisons  fair. For all implementations, trees are grown to full depth, 100 trees are constructed, and $d = \sqrt{p}$ features sampled at each node. For \Sporf,  $\lambda = 1/p$. Experiments are run  using four Intel Xeon E7-4860 v2 processors clocked at 2.60GHz, each processor having 12 physical cores and 24 threads. The amount of available memory is 1 TB DDR3-1600. The operating system is Ubuntu 16.04. Comparisons use three openly available large datasets:

\begin{description}
\item[MNIST]
The MNIST dataset \citep{mnist} has 60,000 training observations and 784 (28x28) features. For a small number of cores, \Sporf~is faster than \Xgboost~but slower than Ranger (Figure \ref{fig:training}(A)). However, when 48 cores are used, \Sporf~is as fast as Ranger and still faster than \Xgboost. 

\item[Higgs]
The Higgs dataset (\url{https://www.kaggle.com/c/higgs-boson}) has 250,000 training observations and 31 features.  \Sporf~is as fast as ranger and faster than \Xgboost~when using 48 cores (Figure \ref{fig:training}(B)). 

\item[p53]
The p53 dataset (\url{https://archive.ics.uci.edu/ml/datasets/p53+Mutants}) has 31,159 training observations and 5,409 features. Figure \ref{fig:training}(C) shows a similar trend as for MNIST.  For this dataset, utilizing additional resources with \Sporf~does not provide as much benefit due to the classification task being too easy (all algorithms achieve perfect classification accuracy)---the trees are shallow, causing the overhead cost of multithreading to outweigh the speed increase as a result of parallelism.
\end{description}

Strong scaling is the relative increase in speed of using multiple cores over that of using a single core. In the ideal case, the use of $N$ cores would produce a factor $N$ speedup. 
\Sporf~has the best strong scaling on MNIST (Figure \ref{fig:training}(D)) and Higgs 
(Figure \ref{fig:training}(E)), while it has strong scaling in between that of Ranger and \Xgboost~on the p53 data set (Figure \ref{fig:training}(F)). This is due to the simplicity of the p53 dataset, as discussed above.

Prediction times can be just as, or even more important than training times in certain applications.  For example,  electron microscopy-based connectomics can acquire multi-petabyte datasets that require classification of each voxel~\cite{Motta2019-tx}. Moreover, recent automatic hyperparameter tuning suites incorporate runtime in their evaluations, which leverage out-of-sample prediction accuracy~\cite{Falkner2018-hw}.  Thus, accelerating prediction times can improve the effectiveness of hyperparameter sweeps. 

\begin{figure*}[ht!]
\centering
\centerline{\includegraphics[width=0.8\textwidth,trim={0in 0.3in 0 0},clip]{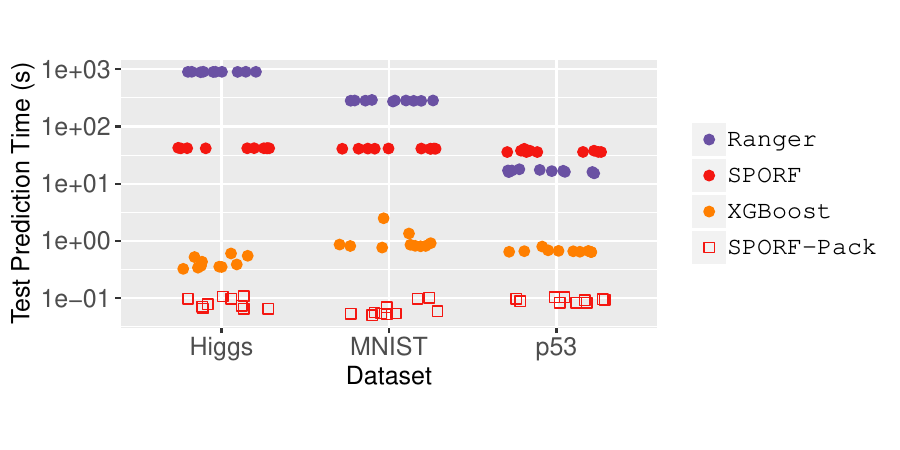}}
\caption{Comparison of test set prediction times.  Forest Packing results show a 10x speed up in real time prediction scenarios.  Test set sizes: Higgs, 50,000 observations; MNIST, 10,000 observations; p53, 6,000 observations.  Predictions were made sequentially without batching.}
\label{fig:forestPacking}
\end{figure*}

Figure \ref{fig:forestPacking} compares the prediction times of the various implementations on the same three datasets. In addition to our standard \Sporf~prediction implementation, we also compare a "Forest Packing" prediction implementation~\cite{Browne2019-sh}. Briefly, Forest Packing is a procedure performed after a forest has been grown that reduces prediction latency by reorganizing and compacting the forest data structure. 
The number of test points used for the Higgs, MNIST, and p53 datasets is 50,000, 10,000, and 6,000, respectively. Predictions were made sequentially without batching using a single core. \Sporf~is significantly faster than Ranger on the Higgs and MNIST datasets, and only marginally slower on the p53 dataset. \Xgboost~is much faster than both \Sporf~and Ranger, which is due to the fact that the \Xgboost~algorithm constructs much shallower trees than the other methods. Most notably, the Forest Packing procedure, which "packs" the trees learned by \Sporf, makes predictions roughly ten times faster than \Xgboost~and over 100 times faster than the standard \Sporf~on all three datasets.

\section{Conclusion}

In this work we showed that existing oblique splitting extensions to \Rf~forfeit some of the nice properties of \Rf, while achieving improved performance in certain settings. We therefore introdced \Sporf~which was designed to preserve the desirable properties of both \Rf~and  oblique forest methods, rendering it statistically robust, computationally efficient, scalable, and interpretable. This work only focused on classification; we also have a preliminary implementation for regression, which seems to perform similarly to \Rf~on a suite of regression benchmark data sets. Future work will investigate the behavior and performance of \Sporf~on univariate and multivariate regression tasks. 

One limitation of using sparse random projections to generate the candidate oblique splits is that it will never find informative splits in cases for which the signal is contained in a dense linear combination of features or nonlinear combinations of features. In such cases, supervised computation of split directions may be more suitable. Perhaps a decision forest method that evaluates both sparse random projections and dense supervised projections at each split node could further improve performance in such settings.

On a more theoretical note, we demonstrated that \Sporf~achieves empirical statistical consistency on a classification problem for which \citet{biau2008} proved that \Rf~cannot achieve better than an error rate of 1/6, even though the Bayes optimal error rate is zero in this setting. This raises a question as to whether it is possible to construct a problem in which \Rf~is consistent and \Sporf~is not. Or could it be the case that \Sporf~is always consistent when \Rf~is? The consistency theorems in \citet{Biau2016} for \Rf~in the case of additive regression models should be extendable to \Sporf~with some minor modifications---their proofs rely on clever adaptations of classical consistency results for data-independent partitioning classifiers, which are agnostic to whether the splits are axis-aligned or not.
Another factor that dictates the lower bound of error rate, as \citet{Breiman2001} proved, is the relative balance between the strength and correlation of trees. Our investigation of strength and correlation on the Sparse Parity, Orthant, and Trunk simulations is offered in Appendix \ref{app:strength}. The results suggest that \Sporf~can outperform other algorithms because of stronger trees and/or less correlated trees. Therefore, \Sporf~perhaps offers more flexible control over the balance between tree strength and correlation, thereby allowing it to adapt better to different problems. 

Our implementation of \Sporf~is as computationally efficient and scalable or more so than existing tree ensemble implementations. Additionally, our implementation can realize many previously proposed tree ensemble methods by allowing the user to define how random projections are generated. Open source code is available at \url{https://neurodata.io/sporf/}, including both the R package discussed here, and a C++ version with both R and Python bindings that we are actively developing. 

  
\section*{Acknowledgments}

This work is graciously supported by the Defense Advanced Research Projects Agency (DARPA) SIMPLEX program through SPAWAR contract N66001-15-C-4041, DARPA GRAPHS N66001-14-1-4028, and DARPA Lifelong Learning Machines program through contract FA8650-18-2-7834.




%

\bibliographystyle{unsrtnat}
\bibliography{RandomerForest}

\begin{thebibliography}{42}
\providecommand{\natexlab}[1]{#1}
\providecommand{\url}[1]{\texttt{#1}}
\expandafter\ifx\csname urlstyle\endcsname\relax
  \providecommand{\doi}[1]{doi: #1}\else
  \providecommand{\doi}{doi: \begingroup \urlstyle{rm}\Url}\fi

\bibitem[Fernandez-Delgado et~al.(2014)Fernandez-Delgado, Cernadas, Barro, and
  Amorim]{Delgado2014}
M.~Fernandez-Delgado, E.~Cernadas, S.~Barro, and D.~Amorim.
\newblock Do we need hundreds of classifiers to solve real world classification
  problems?
\newblock \emph{Journal of Machine Learning Research}, 15\penalty0
  (1):\penalty0 3133--3181, October 2014.

\bibitem[Caruana et~al.(2008)Caruana, Karampatziakis, and
  Yessenalina]{Caruana2008}
R.~Caruana, N.~Karampatziakis, and A.~Yessenalina.
\newblock An empirical evaluation of supervised learning in high dimensions.
\newblock \emph{Proceedings of the 25th International Conference on Machine
  Learning}, 2008.

\bibitem[Caruana and Niculescu-Mizil(2006)]{caruana2006}
Rich Caruana and Alexandru Niculescu-Mizil.
\newblock An empirical comparison of supervised learning algorithms.
\newblock In \emph{Proceedings of the 23rd international conference on Machine
  learning}, pages 161--168. ACM, 2006.

\bibitem[Chen and Guestrin(2016)]{chen2016}
Tianqi Chen and Carlos Guestrin.
\newblock Xgboost: A scalable tree boosting system.
\newblock In \emph{Proceedings of the 22nd acm sigkdd international conference
  on knowledge discovery and data mining}, pages 785--794. ACM, 2016.

\bibitem[Breiman(2001)]{Breiman2001}
L.~Breiman.
\newblock Random forests.
\newblock \emph{Machine Learning}, 4\penalty0 (1):\penalty0 5--32, October
  2001.

\bibitem[Blaser and Fryzlewicz(2016)]{Blaser2016}
Rico Blaser and Piotr Fryzlewicz.
\newblock Random rotation ensembles.
\newblock \emph{Journal of Machine Learning Research}, 17\penalty0
  (4):\penalty0 1--26, 2016.

\bibitem[Rainforth and Wood(2015)]{Rainforth2015}
Tom Rainforth and Frank Wood.
\newblock Canonical correlation forests.
\newblock \emph{arXiv preprint arXiv:1507.05444}, 2015.

\bibitem[Li et~al.(2006)Li, Hastie, and Church]{Li2006}
P.~Li, T.~J. Hastie, and K.~W. Church.
\newblock Very sparse random projections.
\newblock In \emph{Proceedings of the 12th ACM SIGKDD international conference
  on Knowledge discovery and data mining}, pages 287--296. ACM, 2006.

\bibitem[Devroye et~al.(1996)Devroye, Gyorfi, and Lugosi]{DevroyeBook}
L.~Devroye, L.~Gyorfi, and G.~Lugosi.
\newblock \emph{A Probabilistic Theory of Pattern Recognition}.
\newblock Springer Science \& Business Media, 1996.

\bibitem[Tomita et~al.(2017)Tomita, Maggioni, and Vogelstein]{Tomita2017}
Tyler~M Tomita, Mauro Maggioni, and Joshua~T Vogelstein.
\newblock Roflmao: Robust oblique forests with linear matrix operations.
\newblock In \emph{SIAM Data Mining}, 2017.

\bibitem[Heath et~al.(1993)Heath, Kasif, and Salzberg]{Heath1993}
D.~Heath, S.~Kasif, and S.~Salzberg.
\newblock Induction of oblique decision trees.
\newblock \emph{Journal of Artificial Intelligence Research}, 2\penalty0
  (2):\penalty0 1--32, 1993.

\bibitem[Rodriguez et~al.(2006)Rodriguez, Kuncheva, and Alonso]{Rodriguez2006}
J.~J. Rodriguez, L.~I. Kuncheva, and C.~J. Alonso.
\newblock Rotation forest: A new classifier ensemble method.
\newblock \emph{Pattern Analysis and Machine Intelligence, IEEE Transactions
  on}, 28\penalty0 (10):\penalty0 1619--1630, 2006.

\bibitem[Menze et~al.(2011)Menze, Kelm, Splitthoff, Koethe, and
  Hamprecht]{Menze2011}
B.~H. Menze, B.M Kelm, D.~N. Splitthoff, U.~Koethe, and F.~A. Hamprecht.
\newblock On oblique random forests.
\newblock In Dimitrios Gunopulos, Thomas Hofmann, Donato Malerba, and Michalis
  Vazirgiannis, editors, \emph{Machine Learning and Knowledge Discovery in
  Databases}, volume 6912 of \emph{Lecture Notes in Computer Science}, pages
  453--469. Springer Berlin Heidelberg, 2011.
\newblock ISBN 978-3-642-23782-9.
\newblock \doi{10.1007/978-3-642-23783-6_29}.
\newblock URL \url{http://dx.doi.org/10.1007/978-3-642-23783-6_29}.

\bibitem[Lee et~al.(2015)Lee, Yang, and Oh]{lee2015}
Donghoon Lee, Ming-Hsuan Yang, and Songhwai Oh.
\newblock Fast and accurate head pose estimation via random projection forests.
\newblock In \emph{Proceedings of the IEEE International Conference on Computer
  Vision}, pages 1958--1966, 2015.

\bibitem[Bingham and Mannila(2001)]{bingham2001}
Ella Bingham and Heikki Mannila.
\newblock Random projection in dimensionality reduction: applications to image
  and text data.
\newblock In \emph{Proceedings of the seventh ACM SIGKDD international
  conference on Knowledge discovery and data mining}, pages 245--250. ACM,
  2001.

\bibitem[Fern and Brodley(2003)]{fern2003}
Xiaoli~Z Fern and Carla~E Brodley.
\newblock Random projection for high dimensional data clustering: A cluster
  ensemble approach.
\newblock In \emph{Proceedings of the 20th international conference on machine
  learning (ICML-03)}, pages 186--193, 2003.

\bibitem[Fradkin and Madigan(2003)]{fradkin2003}
Dmitriy Fradkin and David Madigan.
\newblock Experiments with random projections for machine learning.
\newblock In \emph{Proceedings of the ninth ACM SIGKDD international conference
  on Knowledge discovery and data mining}, pages 517--522. ACM, 2003.

\bibitem[Achlioptas(2003)]{achlioptas2003}
Dimitris Achlioptas.
\newblock Database-friendly random projections: Johnson-lindenstrauss with
  binary coins.
\newblock \emph{Journal of computer and System Sciences}, 66\penalty0
  (4):\penalty0 671--687, 2003.

\bibitem[Hegde et~al.(2008)Hegde, Wakin, and Baraniuk]{hegde2008}
Chinmay Hegde, Michael Wakin, and Richard Baraniuk.
\newblock Random projections for manifold learning.
\newblock In \emph{Advances in neural information processing systems}, pages
  641--648, 2008.

\bibitem[Dasgupta and Freund(2008)]{Dasgupta2008-rq}
Sanjoy Dasgupta and Yoav Freund.
\newblock Random projection trees and low dimensional manifolds.
\newblock In \emph{Proceedings of the Fortieth Annual {ACM} Symposium on Theory
  of Computing}, STOC '08, pages 537--546, New York, NY, USA, 2008. ACM.

\bibitem[Dasgupta and Freund(2009)]{Dasgupta2009-fa}
S~Dasgupta and Y~Freund.
\newblock Random projection trees for vector quantization.
\newblock \emph{IEEE Transactions on Information}, 2009.

\bibitem[Dasgupta and Sinha(2013)]{Dasgupta2013-sc}
S~Dasgupta and K~Sinha.
\newblock Randomized partition trees for exact nearest neighbor search.
\newblock \emph{Conference on Learning Theory}, 2013.

\bibitem[Breiman et~al.(1998)]{breiman1998}
Leo Breiman et~al.
\newblock Arcing classifier (with discussion and a rejoinder by the author).
\newblock \emph{The annals of statistics}, 26\penalty0 (3):\penalty0 801--849,
  1998.

\bibitem[Friedman(2001)]{friedman2001}
Jerome~H Friedman.
\newblock Greedy function approximation: a gradient boosting machine.
\newblock \emph{Annals of statistics}, pages 1189--1232, 2001.

\bibitem[Wyner et~al.(2017)Wyner, Olson, Bleich, and Mease]{Wyner2017-ax}
A~J Wyner, M~Olson, J~Bleich, and D~Mease.
\newblock Explaining the success of adaboost and random forests as
  interpolating classifiers.
\newblock \emph{J. Mach. Learn. Res.}, 2017.

\bibitem[Ke et~al.(2017)Ke, Meng, Finley, Wang, Chen, Ma, Ye, and Liu]{ke2017}
Guolin Ke, Qi~Meng, Thomas Finley, Taifeng Wang, Wei Chen, Weidong Ma, Qiwei
  Ye, and Tie-Yan Liu.
\newblock Lightgbm: A highly efficient gradient boosting decision tree.
\newblock In I.~Guyon, U.~V. Luxburg, S.~Bengio, H.~Wallach, R.~Fergus,
  S.~Vishwanathan, and R.~Garnett, editors, \emph{Advances in Neural
  Information Processing Systems 30}, pages 3146--3154. Curran Associates,
  Inc., 2017.
\newblock URL
  \url{http://papers.nips.cc/paper/6907-lightgbm-a-highly-efficient-gradient-boosting-decision-tree.pdf}.

\bibitem[Vershynin(2019)]{Vershynin2019}
Roman Vershynin.
\newblock \emph{High Dimensional Probability: An Introduction with Applications
  in Data Science}.
\newblock 2019.
\newblock URL
  \url{https://www.math.uci.edu/~rvershyn/papers/HDP-book/HDP-book.pdf}.

\bibitem[Probst et~al.(2019)Probst, Boulesteix, and Bischl]{Probst2019}
Philipp Probst, Anne-Laure Boulesteix, and Bernd Bischl.
\newblock Tunability: Importance of hyperparameters of machine learning
  algorithms.
\newblock \emph{Journal of Machine Learning Research}, 20\penalty0
  (53):\penalty0 1--32, 2019.

\bibitem[Breiman and Cutler(2002)]{Breiman2002}
Leo Breiman and Adele Cutler.
\newblock \emph{Random Forests}, 2002.
\newblock URL
  \url{https://www.stat.berkeley.edu/~breiman/RandomForests/cc_home.htm}.

\bibitem[Biau et~al.(2008)Biau, Devroye, and Lugosi]{biau2008}
G.~Biau, L.~Devroye, and G.~Lugosi.
\newblock Consistency of random forests and other averaging classifiers.
\newblock \emph{The Journal of Machine Learning Research}, 9:\penalty0
  2015--2033, 2008.

\bibitem[Trunk(1979)]{Trunk1979}
G.~V. Trunk.
\newblock A problem of dimensionality: A simple example.
\newblock \emph{Pattern Analysis and Machine Intelligence, IEEE Transactions
  on}, 1\penalty0 (3):\penalty0 306--307, 1979.

\bibitem[Louppe(2014)]{Louppe2014}
Gilles Louppe.
\newblock Understanding random forests: From theory to practice.
\newblock \emph{arXiv preprint arXiv:1407.7502}, 2014.

\bibitem[Browne et~al.(2018)Browne, Tomita, and Vogelstein]{rerf}
James Browne, Tyler Tomita, and Joshua~T. Vogelstein.
\newblock \emph{rerf: Randomer Forest}, 2018.
\newblock URL \url{https://cran.r-project.org/web/packages/rerf/}.

\bibitem[Eddelbuettel(2018)]{Rcpp}
Dirk Eddelbuettel.
\newblock \emph{Rcpp: Seamless R and C++ Integration}, 2018.
\newblock URL \url{https://cran.r-project.org/web/packages/Rcpp/index.html}.

\bibitem[Chen(2018)]{xgboost}
Tianqi Chen.
\newblock \emph{xgboost: Extreme Gradient Boosting}, 2018.
\newblock URL \url{https://cran.r-project.org/web/packages/xgboost/}.

\bibitem[Wright(2018)]{ranger}
Marvin~N. Wright.
\newblock \emph{ranger: A fast Implementation of Random Forests}, 2018.
\newblock URL \url{https://cran.r-project.org/web/packages/ranger/}.

\bibitem[Lecun et~al.()Lecun, Cortes, and Burges]{mnist}
Yann Lecun, Corinna Cortes, and Christopher~J.C. Burges.
\newblock \emph{The MNIST Database of Handwritten Digits}.
\newblock URL \url{http://yann.lecun.com/exdb/mnist/}.

\bibitem[Motta et~al.(2019)Motta, Schurr, Staffler, and
  Helmstaedter]{Motta2019-tx}
Alessandro Motta, Meike Schurr, Benedikt Staffler, and Moritz Helmstaedter.
\newblock {Big data in nanoscale connectomics, and the greed for training
  labels}.
\newblock \emph{Curr. Opin. Neurobiol.}, 55:\penalty0 180--187, May 2019.

\bibitem[Falkner et~al.(2018)Falkner, Klein, and Hutter]{Falkner2018-hw}
Stefan Falkner, Aaron Klein, and Frank Hutter.
\newblock Bohb: Robust and efficient hyperparameter optimization at scale.
\newblock \emph{arXiv preprint arXiv:1807.01774}, 2018.

\bibitem[Browne et~al.(2019)Browne, Mhembere, Tomita, Vogelstein, and
  Burns]{Browne2019-sh}
J~Browne, D~Mhembere, T~Tomita, J~Vogelstein, and R~Burns.
\newblock {Forest Packing: Fast Parallel, Decision Forests}.
\newblock In \emph{{Proceedings of the 2019 SIAM International Conference on
  Data Mining}}, Proceedings, pages 46--54. Society for Industrial and Applied
  Mathematics, May 2019.

\bibitem[Biau et~al.(2016)Biau, Scornet, and Welbl]{Biau2016}
G{\'e}rard Biau, Erwan Scornet, and Johannes Welbl.
\newblock Neural random forests.
\newblock \emph{arXiv preprint arXiv:1604.07143}, 2016.

\bibitem[James(2003)]{James2003}
Gareth~M James.
\newblock Variance and bias for general loss functions.
\newblock \emph{Machine Learning}, 51\penalty0 (2):\penalty0 115--135, 2003.

\end{thebibliography}


\begin{appendices}

\section{Hyperparameter Tuning}
\label{section: tuning_app}

Hyperparameters in \Xgboost~are tuned via grid search using the R caret package. The values tried for each hyperparameter are based on suggestions by Owen Zhang (\url{https://www.slideshare.net/OwenZhang2/tips-for-data-science-competitions}), a research data scientist who has had many successes in data science competitions using \Xgboost:
\begin{itemize}
\item nrounds: 100, 1000
\item subsample: 0.5, 0.75, 1 
\item eta: 0.001, 0.01
\item colsample\_bytree: 0.4, 0.6, 0.8, 1
\item min\_child\_weight: 1
\item max\_depth:~4, 6, 8, 10, 100000
\item gamma: 0
\end{itemize}

Selection of the hyperparameter \textbf{}values is based on minimization of a five-fold cross-validation error rate.

\section{Real Benchmark datasets}
\label{sec:benchmarks}
We use 105 benchmark datasets from the UCI machine learning repository {for classification}. These datasets are most of the datasets used in \citet{Delgado2014}; some were removed due to licensing or unavailability issues. We noticed certain anomalies in \cite{Delgado2014}'s pre-processed data, so we pre-processed the raw data again as follows.

\begin{enumerate}[nolistsep]
\item \textbf{Remove of nonsensical features}. Some features, such as unique sample identifiers, or features that were the same value for every sample, were removed.
\item \textbf{Impute missing values}. The R randomForest package was used to impute missing values. This method was chosen because it is nonparametric and is one of the few imputation methods that can natively impute missing categorical entries.
\item \textbf{One-hot-encode categorical features}. Most classifiers cannot handle categorical data natively. Given a categorical feature with possible values $\{c_1,\ldots,c_m\}$, we expand to $m$ binary features. If a data point has categorical value $c_k, \forall k \in 1,\ldots,m$ then the $k^{th}$ binary feature is assigned a value of one and zero otherwise.
\item \textbf{Integer encoding of ordinal features}. Categorical features having order to them, such as "cold", "luke-warm", and "hot", were numerically encoded to respect this ordering with integers starting from 1.
\item \textbf{Standardization of the format}. Lastly, all datasets were stored as CSV files, with rows representing observations and columns representing features. The class labels were placed as the last column.
\item \textbf{Five-fold parition}. Each dataset was randomly divided into five partitions for five-fold cross-validation. Partitions preserved the relative class frequencies by stratification. Each partition included a different $20\%$ of the data for testing.
\end{enumerate}





\section{Strength and Correlation of Trees}
\label{app:strength}

\begin{figure*}[ht!]
\centering
\centerline{\includegraphics[width=1\textwidth]{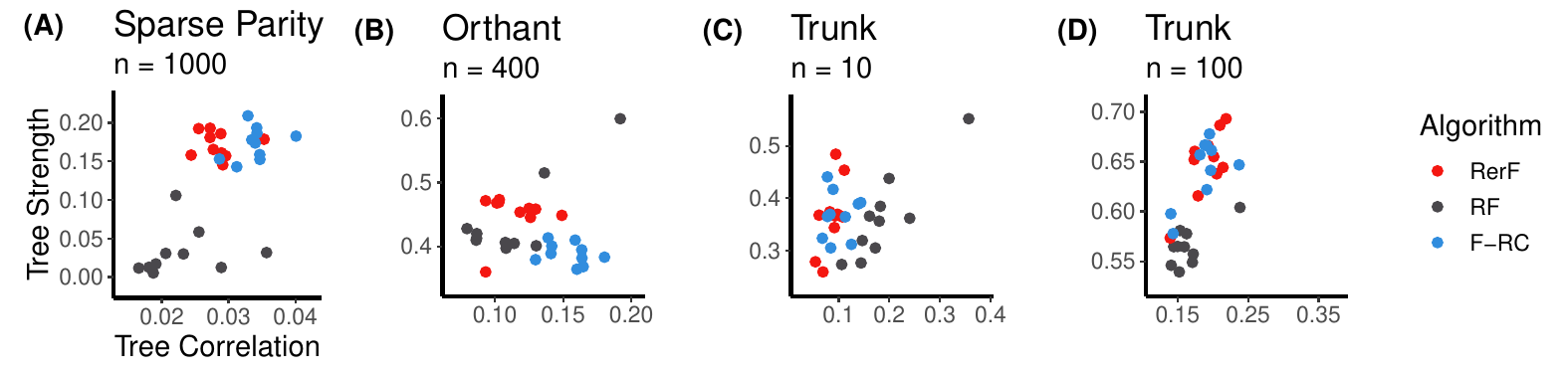}}
\caption{Comparison of tree strength and correlation of \Sporf, \Rf, and \Frc~on four of the simulated datasets: \textbf{(A)} sparse parity with $p = 10, n = 1000$, \textbf{(B)} orthant with $p = 6, n = 400$, \textbf{(C)} Trunk with $p = 10, n = 10$, and \textbf{(D)} Trunk with $p = 10, n = 100$. For a particular algorithm, there are ten dots, each corresponding to one of ten trials. Note in all settings, \Sporf~beats \Rf~and/or \Frc. However, the mechanism by which it does varies across the different settings. In sparse parity \Sporf~wins because the trees are substantially stronger, even though the correlation increases. In Trunk for small sample size, it is purely because of less correlated trees. However, when sample size increases 10-fold, it wins purely because of stronger trees. This suggests that \Sporf~can effectively trade-off strength for correlation on the basis of sample complexity to empirically outperform \Rf~and \Frc.}
\label{fig:strcorr_synthetic}
\end{figure*}

One of the most important and well-known results in ensemble learning theory for classification states that the generalization error of an ensemble learning procedure is bounded above by the quantity ${\bar{\rho} (1 - s^2)}/{s^2}$, where $\bar{\rho}$ is a particular measure of the correlation of the base learners and $s$ is a particular measure of the strength of the base learners \citep{Breiman2001}. In both \Sporf~and \Frc, the set of possible splits that can be sampled is far larger in size than that for \Rf, which may lead to more diverse trees. Moreover, the ability to sample a more diverse set of splits may increase the likelihood of finding good splits and therefore boost the strength of the trees. To investigate the strength and correlation of trees using different projection distributions, we evaluate \Rf, \Frc, and \Sporf~on the three simulation settings described above.
Scatter plots of tree strength vs tree correlation are shown in Figure \ref{fig:strcorr_synthetic} for sparse parity ($n = 1000$), orthant ($n = 400$), Trunk ($n = 10$), and Trunk ($n = 100$). In all four settings, \Sporf~classifies as well as or better than \Rf~and \Frc.

On the sparse parity setting, \Sporf~and \Frc~produce significantly stronger trees than does \Rf, at the expense of an increase in correlation among the trees (Figure \ref{fig:strcorr_synthetic}(A)). Both \Sporf~and \Frc~are much more accurate than \Rf~in this setting, so any performance degradation due to the increase in correlation relative to \Rf~is outweighed by the increased strength. \Sporf~produces slightly less correlated trees than does \Frc, which may explain why \Sporf~has a slightly lower error rate than does \Frc~on this setting.

On the orthant setting, \Frc~produces trees of roughly the same strength as those in \Rf, but significantly more correlated (Figure \ref{fig:strcorr_synthetic}(B)). This may explain why \Frc~has substantially worse prediction accuracy than does \Rf. \Sporf~also produces trees more correlated than those in \Rf, but to a lesser extent than \Frc. Furthermore, the trees in \Sporf~are stronger than those in \Rf. Observing that \Sporf~has roughly the same error rate as \Rf~does, it seems that any contribution of greater tree strength in \Sporf~is canceled by a contribution of greater tree correlation.

On the Trunk setting with $p = 10$ and $n = 10$, \Sporf~and \Frc~produces trees that are comparable in strength to those in \Rf~but less correlated (Figure \ref{fig:strcorr_synthetic}(C)). However, when increasing $n$ to $100$, the trees in \Sporf~and \Frc~become both stronger and more correlated. In both cases, \Sporf~and \Frc~have better classification performance than \Rf.

These results 
suggest a possibly general phenomenon. Namely, for smaller training set sizes, tree correlation may be a more important factor than tree strength because there  is not enough data to induce strong trees, and thus, the only way to improve performance is through increasing the diversity of trees. Likewise, when the training set is sufficiently large, tree correlation matters less because there is enough data to induce strong trees. Since \Sporf~has the ability to produce both stronger and more diverse trees than \Rf, it is adaptive to both regimes In all four settings, \Sporf~never produces more correlated trees than does \Frc, and sometimes produces less correlated trees. A possible explanation for this is that the splits made by \Sporf~are linear combinations of a random number of dimensions, whereas in \Frc~the splits are linear combinations of a fixed number of dimensions. Thus, in some sense, there is more randomness in \Sporf~than in \Frc.

\section{Understanding the Bias and Variance of \Sporf}

The crux of supervised learning tasks is to optimize the trade-off between bias and variance. As a first step in understanding how the choice of projection distribution effects the balance between bias and variance, we estimate bias, variance, and error rate of the various algorithms on the sparse parity problem. Universally agreed upon definitions of bias and variance for 0-1 loss do not exist, and several such definitions have been proposed for each. Here we adopt the framework for defining bias and variance for 0-1 loss proposed in \citet{James2003}. Under this framework, bias and variance for 0-1 loss have similar interpretations to those for mean squared error. That is, bias is a measure of the distance between the expected output of a classifier and the true output, and variance is a measure of the average deviation of a classifier output around its expected output. Unfortunately, these definitions (along with the term for Bayes error) do not provide an additive decomposition for the expected 0-1 loss. Therefore, \citet{James2003} provides two additional statistics that do provide an additive decomposition. In this decomposition, the so-called "systematic effect" measures the contribution of bias to the error rate, while the "variance effect" measures the contribution of variance to the error rate. For completeness, we restate these definitions below.

\begin{figure*}[ht!]
\centering
{\fontfamily{lmss}\selectfont 
\textbf{Sparse Parity}
}
\linebreak
\newline
\centerline{\includegraphics[width=\textwidth]{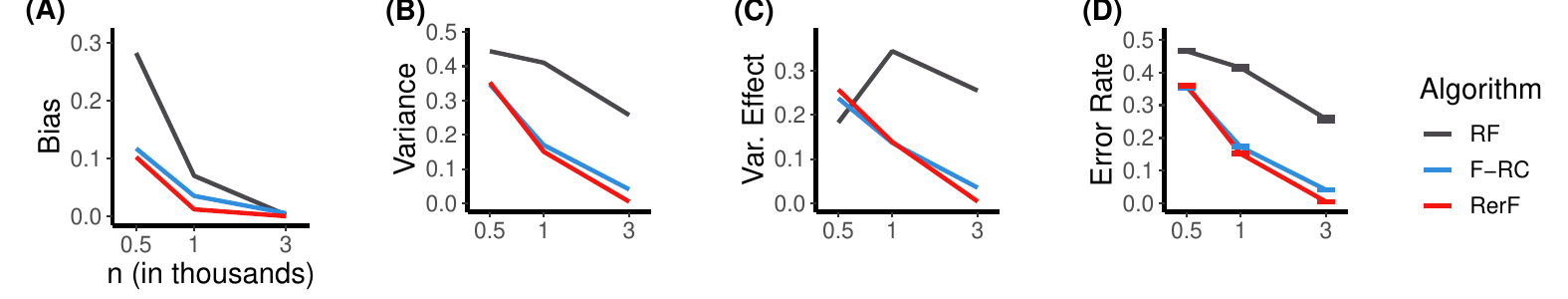}}
\caption{\textbf{(A-D)} Bias, variance, variance effect, and error rate, respectively, on the sparse parity problem as a function of the number of training samples. Error rate is the sum of systematic effect and variance effect, which roughly measure the contributions of bias and variance to the error rate, respectively. In this example, bias and systematic effect are identical because the Bayes error is zero (refer to \citep{James2003}). For smaller training sets, \Sporf~wins primarily through lower bias/systematic effect, while for larger training sets it wins primarily through lower variance effect.}
\label{fig:bv_synthetic}
\end{figure*}

Let $\bar{h}(X) = \argmax\limits_k P_{\mc{D}_n}(h(X | \mc{D}_n) = k)$ be the most common prediction (mode) with respect to the distribution of $\mc{D}_n$. This is referred to as the "systematic" prediction in \citet{James2003}. Furthermore, let $P^*(X) = P_{Y|X}(Y = h^*(X) | X)$ and $\bar{P}(X) = P_{\mc{D}_n}(h(X | \mc{D}_n) = \bar{h}(X))$. The bias, variance, systematic effect (SE), and variance effect (VE) are defined as
\begin{align*}
Bias &= P_X(\bar{h}(X) = h^*(X)), \\
Var &= 1 - E_X[\bar{P}(X)], \\
SE &= E_X[P^*(X) - P_{Y|X}(Y = \bar{h}(X) | X)],\\
VE &= E_X[P_{Y|X}(Y = \bar{h}(X) | X)\\
&- \sum\limits_k P_{Y|X}(Y = k | X)P_{\mc{D}_n}(h(X | \mc{D}_n) = k)].
\end{align*}
Figure \ref{fig:bv_synthetic} compares estimates of  bias, variance, variance effect, and error rate for \Sporf, \Rf, and \Frc~as a function of number of training samples. Since the Bayes error is zero in these settings, systematic effect is the same as bias. The four metrics are estimated from 100 repeated experiments for each value of $n$. In Figure \ref{fig:bv_synthetic}(A), \Sporf~has lower bias than both \Rf~and \Frc~for all training set sizes. All algorithms converge to approximately zero bias after about 3000 samples. Figure \ref{fig:bv_synthetic}(B) shows that \Rf~has substantially more variance than do \Sporf~and \Frc, and \Sporf~has slightly less variance than \Frc~at 3,000 samples. The trend in Figure \ref{fig:bv_synthetic}(C) is similar to that in Figure \ref{fig:bv_synthetic}(B), which is not too surprising since VE measures the contribution of the variance to the error rate. Interestingly, although \Rf~has noticeably more variance at 500 samples than do \Sporf~and \Frc, it has slightly lower VE. It is also surprising that the VE of \Rf~increases from 500 to 1000 training samples. It could be that this is the result of the tradeoff of the substantial reduction in bias. In Figure \ref{fig:bv_synthetic}(D), the error rate is shown for reference, which is the sum of bias and VE. Overall, these results suggest that \Sporf~wins on the sparse parity problem with a small sample size
primarily through lower bias/SE, 
while with a larger sample size it wins mainly via lower variance/VE. 
A similar trend holds for the orthant problem (not shown).

\section{Algorithms}

\begin{algorithm}[!ht]
  \caption{Learning a \Sporf~classification tree. 
}
  \label{alg:rerftrain}
\begin{algorithmic}[1]
  \Require (1) $\mathcal{D}_n = (\mathbf{X},\mathbf{y}) \in \Real^{n \times p} \times \mathcal{Y}^n$: training data (2) $\Theta$: set of split eligibility criteria
  \Ensure A \Sporf~decision tree $T$
  \Function{$T =$ growtree}{$\mathbf{X},\mathbf{y},\Theta$}
  \State $c = 1$
  \Comment{$c$ is the current node index}
  \State $M = 1$
  \Comment $M$ is the number of nodes currently existing
  \State $S^{(c)} =$ bootstrap($\{1,...,n\}$)
  \Comment $S^{(c)}$ is the indices of the observations at node $c$
  \While{$c < M + 1$}
  \Comment visit each of the existing nodes
  \State $(\mathbf{X'},\mathbf{y'}) = (\mathbf{x}_i,y_i)_{i \in S^{(c)}}$
  \Comment data at the current node
  \Linefor{$k = 1,\ldots,K$}{$n_k^{(c)} = \sum_{i \in S^{(c)}} I[y_i = k]$}
  \Comment class counts
  \If{$\Theta$ satisfied}
  \Comment do we split this node?
  \State $\mathbf{A} = [\mathbf{a}_1 \cdots \mathbf{a}_d] \sim f_{\mathbf{A}}$
  \Comment sample random $p \times d$ matrix as defined in \ref{section: sporf}
  \State $\mathbf{\widetilde{X}} = \mathbf{X'}\mathbf{A} = (\mathbf{\widetilde{x}}_i)_{i \in S^{(c)}}$
  \Comment random projection into new feature space
  \State $(j^*,t^*) =$ findbestsplit($\mathbf{\widetilde{X}},\mathbf{y'}$)
  \Comment Algorithm \ref{alg:rerfsplit} 
  \State $S^{(M+1)} = \{i: \mathbf{\widetilde{x}}_i \cdot \mathbf{a}_{j^*} \leq t^* \quad \forall i \in S^{(c)}\}$
  \Comment assign to left child node
  \State $S^{(M+2)} = \{i: \mathbf{\widetilde{x}}_i \cdot \mathbf{a}_{j^*} > t^* \quad \forall i \in S^{(c)}\}$
  \Comment assign to right child node
  \State $\mathbf{a}^{*(c)} = \mathbf{a}_{j^*}$
  \Comment store best projection for current node
  \State $\tau^{*(c)} = t^*$
  \Comment store best split threshold for current node
  \State $\kappa^{(c)} = \{M+1,M + 2\}$
  \Comment node indices of children of current node
  \State $M = M + 2$
  \Comment update the number of nodes that exist
  \Else
  \State $(\mathbf{a}^{*(c)},\tau^{*(c)},\kappa^{*(c)}) =$ NULL
  \EndIf
  \State $c = c + 1$
  \Comment move to next node
  \EndWhile
  \State \Return $(S^{(1)},\{\mathbf{a}^{*(c)},\tau^{*(c)},\kappa^{(c)},\{n_k^{(c)}\}_{k \in \mathcal{Y}}\}_{c=1}^{m-1})$
  \EndFunction
\end{algorithmic}
\end{algorithm}

\begin{algorithm}[!ht]
  \caption{Finding the best node split. This function is called by growtree (Alg \ref{alg:rerftrain}) at every split node. For each of the $p$ dimensions in $\mathbf{X} \in \Real^{n \times p}$, a binary split is assessed at each location between adjacent observations. The dimension $j^*$ and split value $\tau^*$ in $j^*$ that best split the data are selected. The notion of ``best'' means maximizing some choice in scoring function. In classification, the scoring function is typically the reduction in Gini impurity or entropy. 
  The increment function called within this function updates the counts in the left and right partitions as the split is incrementally moved to the right.}
  \label{alg:rerfsplit}
\begin{algorithmic}[1]
  \Require (1) $(\mathbf{X},\mathbf{y}) \in \Real^{n \times p} \times \mathcal{Y}^n$, where $ \mathcal{Y} = \{1,\ldots,K\}$
  \Ensure (1) dimension $j^*$, (2) split value $\tau^*$
  \Function{$(j^*,\tau^*) =$ findbestsplit}{$\mathbf{X},\mathbf{y}$}
  \For{$j = 1,\ldots,p$}
  \State Let $\mathbf{x}^{(j)} = (x_1^{(j)},\ldots,x_n^{(j)})^T$ be the $jth$ column of $\mathbf{X}$.
  \State $\{m_i^j\}_{i \in [n]} =$ sort($\mathbf{x}^{(j)}$)
  \Comment $m_i^j$ is the index of the $i^{th}$ smallest value in $\mathbf{x}^{(j)}$
  \State $t = 0$
  \Comment initialize split to the left of all observations
  \State $n' = 0$
  \Comment number of observations left of the current split
  \State $n'' = n$
  \Comment number of observations right of the current split
  \For{$k = 1,\ldots,K$}
  \State $n_k = \sum_{i=1}^n I[y_i = k]$
  \Comment total number of observations in class $k$
  \State $n'_k = 0$
  \Comment number of observations in class $k$ left of the current split
  \State $n''_k = n_k$
  \Comment number of observations in class $k$ right of the current split
  \EndFor
  \For{$t = 1,\ldots,n-1$}
  \Comment assess split location, moving right one at a time
  \State $(\{(n'_k,n''_k)\},n',n'',y_{m_t^j}) =$ increment($\{(n'_k,n''_k)\},n',n'',y_{m_t^j}$)
  \State $Q^{(j,t)} =$ score($\{(n'_k,n''_k)\},n',n''$)
  \Comment measure of split quality
  \EndFor
  \EndFor
  \State $(j^*,t^*) = \argmax\limits_{j,t} Q^{(j,t)}$
  \Linefor{$i = 0,1$}{$c_i = m_{t^* + i}^{j^*}$}
  \State $\tau^* = \frac{1}{2}(x_{c_0}^{(j^*)} + x_{c_1}^{(j^*)})$
  \Comment compute the actual split location from the index $j^*$
  \State \Return $(j^*,\tau^*)$
  \EndFunction
\end{algorithmic}
\end{algorithm}

\clearpage

\section{Data Tables}
\label{app: tables}
\setlength\LTleft{-0.5in}
\setlength\LTright{-0.5in}
\begin{landscape}
\begin{longtable}{r|llll|lllll}
\caption{Five-fold cross-validation Cohen's kappa values (mean$\pm$SEM) on the UCI datasets, along with summary statistics for each dataset. $n$ is the number of examples, $p_{num}$ is the number of numeric features, $p_{cat}$ is the number of categorical features, and $C$ is the number of classes. Best performing algorithm for each data set is highlighted in bold text.} \label{tab:class} \\
  \hline
 & & & & & \multicolumn{5}{c}{5-fold CV Error Rate} \\
Dataset & $n$ & $p_{num}$ & $p_{cat}$ & $C$ & \Sporf & \Rf & \Xgboost & \Rrrf & \Ccf \\ 
  \hline
abalone & 4177 & 7 & 1 & 28 & $11.2 \pm 0.8$ & $11 \pm 0.8$ & $\mathbf{11.8 \pm 0.4}$ & $9.1 \pm 0.5$ & $10.4 \pm 0.7$ \\ 
  acute\_inflammation\_task\_1 & 120 & 6 & 0 & 2 & $\mathbf{100 \pm 0}$ & $\mathbf{100 \pm 0}$ & $95 \pm 2$ & $\mathbf{100 \pm 0}$ & $\mathbf{100 \pm 0}$ \\ 
  acute\_inflammation\_task\_2 & 120 & 6 & 0 & 2 & $\mathbf{100 \pm 0}$ & $\mathbf{100 \pm 0}$ & $87 \pm 9$ & $\mathbf{100 \pm 0}$ & $\mathbf{100 \pm 0}$ \\ 
  adult & 32561 & 7 & 7 & 2 & $82.38 \pm 0.29$ & $82.42 \pm 0.28$ & $\mathbf{83.45 \pm 0.28}$ & $78.16 \pm 0.39$ & $80.2 \pm 0.23$ \\ 
  annealing & 798 & 27 & 5 & 5 & $\mathbf{98 \pm 1}$ & $\mathbf{98 \pm 1}$ & $97 \pm 1$ & $91 \pm 2$ & $97 \pm 1$ \\ 
  arrhythmia & 452 & 279 & 0 & 13 & $72 \pm 2$ & $\mathbf{74 \pm 2}$ & $72 \pm 2$ & $61 \pm 2$ & $66 \pm 1$ \\ 
  audiology\_std & 200 & 68 & 1 & 24 & $75 \pm 4$ & $73 \pm 5$ & $73 \pm 5$ & $65 \pm 4$ & $\mathbf{78 \pm 5}$ \\ 
  balance\_scale & 625 & 4 & 0 & 3 & $\mathbf{94 \pm 1}$ & $77 \pm 3$ & $83 \pm 1$ & $82 \pm 1$ & $90 \pm 1$ \\ 
  balloons & 16 & 4 & 0 & 2 & $40 \pm 20$ & $50 \pm 20$ & $10 \pm 30$ & $\mathbf{60 \pm 10}$ & $40 \pm 20$ \\ 
  bank & 4521 & 11 & 5 & 2 & $91.7 \pm 0.1$ & $\mathbf{91.9 \pm 0.1}$ & $\mathbf{91.9 \pm 0.1}$ & $91.3 \pm 0.4$ & $91.6 \pm 0.3$ \\ 
  blood & 748 & 4 & 0 & 2 & $71 \pm 1$ & $\mathbf{72 \pm 1}$ & $\mathbf{72 \pm 1}$ & $71 \pm 1$ & $70 \pm 1$ \\ 
  breast\_cancer & 286 & 7 & 2 & 2 & $\mathbf{61 \pm 2}$ & $60 \pm 3$ & $58 \pm 1$ & $54 \pm 4$ & $57 \pm 3$ \\ 
  breast\_cancer-wisconsin & 699 & 9 & 0 & 2 & $\mathbf{96 \pm 2}$ & $\mathbf{96 \pm 2}$ & $95 \pm 2$ & $\mathbf{96 \pm 2}$ & $\mathbf{96 \pm 2}$ \\ 
  breast\_cancer-wisconsin-diag & 569 & 30 & 0 & 2 & $96 \pm 1$ & $94 \pm 1$ & $94 \pm 1$ & $96 \pm 1$ & $\mathbf{97 \pm 1}$ \\ 
  breast\_cancer-wisconsin-prog & 198 & 33 & 0 & 2 & $\mathbf{73 \pm 2}$ & $69 \pm 3$ & $72 \pm 2$ & $72 \pm 2$ & $72 \pm 3$ \\ 
  car & 1728 & 6 & 0 & 4 & $96.5 \pm 0.2$ & $93.1 \pm 0.8$ & $96.5 \pm 0.4$ & $81.5 \pm 1.4$ & $\mathbf{96.9 \pm 0.6}$ \\ 
  cardiotocography\_task\_1 & 2126 & 21 & 0 & 10 & $83.5 \pm 0.5$ & $82.5 \pm 0.6$ & $\mathbf{84.4 \pm 0.5}$ & $74.7 \pm 0.8$ & $81.3 \pm 1.1$ \\ 
  cardiotocography\_task\_2 & 2126 & 21 & 0 & 3 & $93.9 \pm 0.5$ & $93.6 \pm 0.6$ & $\mathbf{94.5 \pm 0.4}$ & $90.1 \pm 0.8$ & $92.2 \pm 0.5$ \\ 
  chess\_krvk & 28056 & 0 & 6 & 18 & $84.01 \pm 0.18$ & $77.99 \pm 0.18$ & $\mathbf{86.76 \pm 0.35}$ & $59.17 \pm 0.13$ & $82.62 \pm 0.22$ \\ 
  chess\_krvkp & 3196 & 35 & 1 & 2 & $\mathbf{99.1 \pm 0.2}$ & $98.8 \pm 0.2$ & $98.8 \pm 0.3$ & $95.7 \pm 0.6$ & $98.8 \pm 0.1$ \\ 
  congressional\_voting & 435 & 16 & 0 & 2 & $94 \pm 2$ & $94 \pm 2$ & $\mathbf{96 \pm 1}$ & $94 \pm 1$ & $95 \pm 2$ \\ 
  conn\_bench-sonar-mines-rocks & 208 & 60 & 0 & 2 & $72 \pm 4$ & $72 \pm 6$ & $\mathbf{77 \pm 7}$ & $70 \pm 3$ & $75 \pm 5$ \\ 
  conn\_bench-vowel-deterding & 528 & 11 & 0 & 11 & $97 \pm 0$ & $96 \pm 1$ & $90 \pm 2$ & $97 \pm 0$ & $\mathbf{98 \pm 1}$ \\ 
  contrac & 1473 & 8 & 1 & 3 & $29.3 \pm 2.8$ & $26.5 \pm 1.9$ & $\mathbf{31.6 \pm 1.7}$ & $24.9 \pm 1.4$ & $28 \pm 1.5$ \\ 
  credit\_approval & 690 & 10 & 5 & 2 & $77 \pm 1$ & $\mathbf{78 \pm 1}$ & $77 \pm 1$ & $74 \pm 2$ & $75 \pm 2$ \\ 
  dermatology & 366 & 34 & 0 & 6 & $\mathbf{98 \pm 1}$ & $\mathbf{98 \pm 1}$ & $97 \pm 1$ & $96 \pm 0$ & $96 \pm 1$ \\ 
  ecoli & 336 & 7 & 0 & 8 & $\mathbf{83 \pm 1}$ & $81 \pm 2$ & $81 \pm 1$ & $\mathbf{83 \pm 1}$ & $81 \pm 1$ \\ 
  flags & 194 & 22 & 6 & 8 & $57 \pm 3$ & $\mathbf{58 \pm 3}$ & $57 \pm 4$ & $46 \pm 5$ & $54 \pm 2$ \\ 
  glass & 214 & 9 & 0 & 6 & $64 \pm 5$ & $\mathbf{67 \pm 6}$ & $65 \pm 5$ & $56 \pm 7$ & $66 \pm 7$ \\ 
  haberman\_survival & 306 & 3 & 0 & 2 & $\mathbf{63 \pm 3}$ & $61 \pm 3$ & $\mathbf{63 \pm 2}$ & $57 \pm 3$ & $54 \pm 4$ \\ 
  hayes\_roth & 132 & 0 & 4 & 3 & $68 \pm 7$ & $68 \pm 7$ & $64 \pm 8$ & $66 \pm 7$ & $\mathbf{69 \pm 7}$ \\ 
  heart\_cleveland & 303 & 10 & 3 & 5 & $47 \pm 2$ & $48 \pm 2$ & $47 \pm 1$ & $\mathbf{51 \pm 1}$ & $45 \pm 2$ \\ 
  heart\_hungarian & 294 & 10 & 3 & 2 & $\mathbf{89 \pm 3}$ & $87 \pm 2$ & $87 \pm 3$ & $81 \pm 3$ & $85 \pm 4$ \\ 
  heart\_switzerland & 123 & 10 & 3 & 5 & $-5 \pm 6$ & $1 \pm 4$ & $-5 \pm 6$ & $2 \pm 10$ & $\mathbf{6 \pm 10}$ \\ 
  heart\_va & 200 & 10 & 3 & 5 & $13 \pm 4$ & $13 \pm 6$ & $15 \pm 5$ & $\mathbf{17 \pm 5}$ & $15 \pm 3$ \\ 
  hepatitis & 155 & 19 & 0 & 2 & $44 \pm 7$ & $42 \pm 15$ & $37 \pm 7$ & $38 \pm 7$ & $\mathbf{54 \pm 10}$ \\ 
  hill\_valley & 606 & 100 & 0 & 2 & $\mathbf{100 \pm 0}$ & $12 \pm 2$ & $22 \pm 4$ & $88 \pm 2$ & $\mathbf{100 \pm 0}$ \\ 
  hill\_valley-noise & 606 & 100 & 0 & 2 & $\mathbf{90 \pm 3}$ & $3 \pm 6$ & $6 \pm 3$ & $65 \pm 2$ & $89 \pm 2$ \\ 
  horse\_colic & 300 & 17 & 4 & 2 & $76 \pm 3$ & $74 \pm 4$ & $\mathbf{80 \pm 1}$ & $77 \pm 3$ & $77 \pm 3$ \\ 
  ilpd\_indian-liver & 583 & 10 & 0 & 2 & $60 \pm 2$ & $59 \pm 2$ & $60 \pm 1$ & $62 \pm 3$ & $\mathbf{63 \pm 1}$ \\ 
  image\_segmentation & 210 & 19 & 0 & 7 & $\mathbf{93 \pm 3}$ & $92 \pm 3$ & $91 \pm 2$ & $87 \pm 3$ & $\mathbf{93 \pm 3}$ \\ 
  ionosphere & 351 & 34 & 0 & 2 & $85 \pm 2$ & $82 \pm 2$ & $81 \pm 1$ & $\mathbf{88 \pm 1}$ & $86 \pm 2$ \\ 
  iris & 150 & 4 & 0 & 3 & $91 \pm 2$ & $94 \pm 3$ & $92 \pm 2$ & $94 \pm 2$ & $\mathbf{96 \pm 2}$ \\ 
  led\_display & 1000 & 7 & 0 & 10 & $68.2 \pm 1.3$ & $68.6 \pm 1.6$ & $\mathbf{69.5 \pm 1.2}$ & $67.9 \pm 1.4$ & $67.6 \pm 1.3$ \\ 
  lenses & 24 & 4 & 0 & 3 & $50 \pm 20$ & $40 \pm 20$ & $\mathbf{60 \pm 20}$ & $30 \pm 10$ & $40 \pm 20$ \\ 
  letter & 20000 & 16 & 0 & 26 & $96.85 \pm 0.13$ & $96.37 \pm 0.11$ & $96.32 \pm 0.05$ & $95.24 \pm 0.22$ & $\mathbf{97.67 \pm 0.18}$ \\ 
  libras & 360 & 90 & 0 & 15 & $85 \pm 2$ & $80 \pm 2$ & $76 \pm 2$ & $84 \pm 2$ & $\mathbf{90 \pm 2}$ \\ 
  low\_res-spect & 531 & 100 & 1 & 48 & $59 \pm 3$ & $51 \pm 2$ & $48 \pm 3$ & $48 \pm 1$ & $\mathbf{62 \pm 1}$ \\ 
  lung\_cancer & 32 & 13 & 43 & 3 & $30 \pm 10$ & $\mathbf{40 \pm 10}$ & $30 \pm 20$ & $20 \pm 10$ & $0 \pm 10$ \\ 
  magic & 19020 & 10 & 0 & 2 & $\mathbf{82.65 \pm 0.3}$ & $81.48 \pm 0.39$ & $82.35 \pm 0.24$ & $79.55 \pm 0.22$ & $81.92 \pm 0.37$ \\ 
  mammographic & 961 & 3 & 2 & 2 & $\mathbf{69 \pm 2}$ & $\mathbf{69 \pm 1}$ & $\mathbf{69 \pm 1}$ & $57 \pm 1$ & $61 \pm 2$ \\ 
  molec\_biol-promoter & 106 & 0 & 57 & 4 & $\mathbf{40 \pm 2}$ & $36 \pm 5$ & $32 \pm 2$ & $15 \pm 4$ & $19 \pm 7$ \\ 
  molec\_biol-splice & 3190 & 0 & 60 & 3 & $93 \pm 0.7$ & $93.2 \pm 0.7$ & $\mathbf{94.2 \pm 0.5}$ & $68.9 \pm 0.6$ & $92.8 \pm 0.9$ \\ 
  monks\_1 & 124 & 2 & 4 & 2 & $\mathbf{98 \pm 2}$ & $\mathbf{98 \pm 2}$ & $82 \pm 3$ & $70 \pm 5$ & $81 \pm 5$ \\ 
  monks\_2 & 169 & 2 & 4 & 2 & $37 \pm 5$ & $37 \pm 5$ & $45 \pm 6$ & $30 \pm 3$ & $\mathbf{61 \pm 4}$ \\ 
  monks\_3 & 122 & 2 & 4 & 2 & $86 \pm 4$ & $86 \pm 4$ & $81 \pm 3$ & $\mathbf{87 \pm 5}$ & $81 \pm 4$ \\ 
  mushroom & 8124 & 7 & 15 & 2 & $\mathbf{100 \pm 0}$ & $\mathbf{100 \pm 0}$ & $99.8 \pm 0.1$ & $99.9 \pm 0$ & $\mathbf{100 \pm 0}$ \\ 
  musk\_1 & 476 & 166 & 0 & 2 & $80 \pm 2$ & $79 \pm 4$ & $80 \pm 3$ & $79 \pm 2$ & $\mathbf{83 \pm 2}$ \\ 
  musk\_2 & 6598 & 166 & 0 & 2 & $97.4 \pm 0.3$ & $97.3 \pm 0.4$ & $\mathbf{98.2 \pm 0.2}$ & $95.1 \pm 0.5$ & $97.7 \pm 0.4$ \\ 
  nursery & 12960 & 6 & 2 & 5 & $\mathbf{99.97 \pm 0.02}$ & $99.71 \pm 0.05$ & $99.91 \pm 0.05$ & $96.2 \pm 0.08$ & $99.92 \pm 0.04$ \\ 
  optical & 3823 & 64 & 0 & 10 & $98.1 \pm 0.2$ & $97.9 \pm 0.3$ & $97.8 \pm 0.3$ & $97.7 \pm 0.2$ & $\mathbf{98.6 \pm 0.1}$ \\ 
  ozone & 2534 & 72 & 0 & 2 & $94 \pm 0.3$ & $94.1 \pm 0.3$ & $\mathbf{94.4 \pm 0.3}$ & $93.9 \pm 0.1$ & $94.3 \pm 0.2$ \\ 
  page\_blocks & 5473 & 10 & 0 & 5 & $\mathbf{97.3 \pm 0.2}$ & $97.2 \pm 0.1$ & $\mathbf{97.3 \pm 0.2}$ & $96.9 \pm 0.1$ & $\mathbf{97.3 \pm 0.2}$ \\ 
  parkinsons & 195 & 22 & 0 & 2 & $69 \pm 8$ & $\mathbf{75 \pm 3}$ & $67 \pm 9$ & $67 \pm 10$ & $\mathbf{75 \pm 5}$ \\ 
  pendigits & 7494 & 16 & 0 & 10 & $99.5 \pm 0.1$ & $99.1 \pm 0.1$ & $99.1 \pm 0.1$ & $99.3 \pm 0.1$ & $\mathbf{99.6 \pm 0.1}$ \\ 
  pima & 768 & 8 & 0 & 2 & $\mathbf{66 \pm 4}$ & $64 \pm 4$ & $63 \pm 5$ & $65 \pm 2$ & $64 \pm 3$ \\ 
  pittsburgh\_bridges-MATERIAL & 106 & 4 & 3 & 3 & $\mathbf{55 \pm 5}$ & $\mathbf{55 \pm 5}$ & $51 \pm 5$ & $12 \pm 5$ & $38 \pm 10$ \\ 
  pittsburgh\_bridges-REL-L & 103 & 4 & 3 & 3 & $58 \pm 6$ & $58 \pm 7$ & $59 \pm 7$ & $52 \pm 3$ & $\mathbf{62 \pm 4}$ \\ 
  pittsburgh\_bridges-SPAN & 92 & 4 & 3 & 3 & $40 \pm 10$ & $40 \pm 10$ & $40 \pm 10$ & $40 \pm 10$ & $40 \pm 10$ \\ 
  pittsburgh\_bridges-T-OR-D & 102 & 4 & 3 & 2 & $7 \pm 7$ & $7 \pm 12$ & $\mathbf{33 \pm 11}$ & $7 \pm 7$ & $27 \pm 12$ \\ 
  pittsburgh\_bridges-TYPE & 106 & 4 & 3 & 7 & $\mathbf{39 \pm 4}$ & $37 \pm 4$ & $27 \pm 8$ & $11 \pm 7$ & $24 \pm 5$ \\ 
  planning & 182 & 12 & 0 & 2 & $60 \pm 2$ & $60 \pm 4$ & $48 \pm 5$ & $\mathbf{62 \pm 2}$ & $59 \pm 4$ \\ 
  post\_operative & 90 & 8 & 0 & 3 & $\mathbf{60 \pm 0}$ & $50 \pm 10$ & $\mathbf{60 \pm 0}$ & $50 \pm 0$ & $50 \pm 10$ \\ 
  ringnorm & 7400 & 20 & 0 & 2 & $96.1 \pm 0.2$ & $92.1 \pm 0.5$ & $\mathbf{96.3 \pm 0.4}$ & $95.9 \pm 0.1$ & $95.6 \pm 0.3$ \\ 
  seeds & 210 & 7 & 0 & 3 & $\mathbf{91 \pm 3}$ & $90 \pm 4$ & $89 \pm 4$ & $88 \pm 4$ & $90 \pm 3$ \\ 
  semeion & 1593 & 256 & 0 & 10 & $93.4 \pm 0.4$ & $93.7 \pm 0.7$ & $93.7 \pm 0.5$ & $91 \pm 0.9$ & $\mathbf{94.2 \pm 0.4}$ \\ 
  soybean & 307 & 22 & 13 & 19 & $90 \pm 1$ & $90 \pm 2$ & $90 \pm 2$ & $90 \pm 1$ & $\mathbf{92 \pm 2}$ \\ 
  spambase & 4601 & 57 & 0 & 2 & $92.9 \pm 0.7$ & $92.2 \pm 0.6$ & $92.6 \pm 0.5$ & $90.4 \pm 0.5$ & $\mathbf{93.3 \pm 0.7}$ \\ 
  spect & 80 & 22 & 0 & 2 & $30 \pm 10$ & $\mathbf{40 \pm 10}$ & $\mathbf{40 \pm 10}$ & $\mathbf{40 \pm 20}$ & $30 \pm 10$ \\ 
  spectf & 80 & 44 & 0 & 2 & $\mathbf{60 \pm 10}$ & $50 \pm 10$ & $40 \pm 10$ & $\mathbf{60 \pm 10}$ & $50 \pm 10$ \\ 
  statlog\_australian-credit & 690 & 10 & 4 & 2 & $77 \pm 2$ & $\mathbf{78 \pm 2}$ & $76 \pm 3$ & $72 \pm 1$ & $74 \pm 1$ \\ 
  statlog\_german-credit & 1000 & 14 & 6 & 2 & $\mathbf{66.4 \pm 1.5}$ & $65 \pm 1.7$ & $63.6 \pm 2.9$ & $61.6 \pm 1.7$ & $64.3 \pm 1.6$ \\ 
  statlog\_heart & 270 & 10 & 3 & 2 & $68 \pm 1$ & $70 \pm 2$ & $\mathbf{71 \pm 4}$ & $69 \pm 3$ & $67 \pm 2$ \\ 
  statlog\_image & 2310 & 19 & 0 & 7 & $98 \pm 0.5$ & $97.7 \pm 0.4$ & $98.3 \pm 0.4$ & $96.5 \pm 0.5$ & $\mathbf{98.2 \pm 0.4}$ \\ 
  statlog\_landsat & 4435 & 36 & 0 & 6 & $88.5 \pm 0.5$ & $88.3 \pm 0.6$ & $\mathbf{89.2 \pm 0.5}$ & $87.8 \pm 0.4$ & $88.9 \pm 0.6$ \\ 
  statlog\_shuttle & 43500 & 9 & 0 & 7 & $\mathbf{99.98 \pm 0.01}$ & $99.97 \pm 0.01$ & $99.97 \pm 0.01$ & $99.87 \pm 0.01$ & $99.97 \pm 0.01$ \\ 
  statlog\_vehicle & 846 & 18 & 0 & 4 & $74 \pm 1$ & $68 \pm 2$ & $69 \pm 1$ & $69 \pm 2$ & $\mathbf{77 \pm 0}$ \\ 
  steel\_plates & 1941 & 27 & 0 & 7 & $68.1 \pm 1$ & $69.4 \pm 0.6$ & $\mathbf{71.1 \pm 0.9}$ & $64.4 \pm 1.7$ & $66.4 \pm 1.5$ \\ 
  synthetic\_control & 600 & 60 & 0 & 6 & $98 \pm 1$ & $\mathbf{99 \pm 1}$ & $98 \pm 1$ & $98 \pm 1$ & $\mathbf{99 \pm 0}$ \\ 
  teaching & 151 & 3 & 2 & 3 & $\mathbf{39 \pm 6}$ & $36 \pm 4$ & $30 \pm 3$ & $\mathbf{39 \pm 8}$ & $38 \pm 8$ \\ 
  thyroid & 3772 & 21 & 0 & 3 & $96.8 \pm 1$ & $\mathbf{97.2 \pm 1.4}$ & $96.5 \pm 1.6$ & $38.4 \pm 1.8$ & $93.7 \pm 1.9$ \\ 
  tic\_tac-toe & 958 & 0 & 9 & 2 & $96 \pm 1$ & $\mathbf{97 \pm 1}$ & $\mathbf{97 \pm 1}$ & $55 \pm 3$ & $95 \pm 1$ \\ 
  titanic & 2201 & 2 & 1 & 2 & $\mathbf{69.1 \pm 0.5}$ & $\mathbf{69.1 \pm 0.5}$ & $68.5 \pm 0.4$ & $68.5 \pm 0.4$ & $68.5 \pm 0.4$ \\ 
  twonorm & 7400 & 20 & 0 & 2 & $95.5 \pm 0.3$ & $94.8 \pm 0.2$ & $94.9 \pm 0.3$ & $\mathbf{95.7 \pm 0.3}$ & $\mathbf{95.7 \pm 0.3}$ \\ 
  vertebral\_column\_task\_1 & 310 & 6 & 0 & 2 & $\mathbf{78 \pm 2}$ & $75 \pm 2$ & $72 \pm 1$ & $\mathbf{78 \pm 2}$ & $75 \pm 1$ \\ 
  vertebral\_column\_task\_2 & 310 & 6 & 0 & 3 & $\mathbf{68 \pm 5}$ & $66 \pm 5$ & $66 \pm 4$ & $63 \pm 4$ & $67 \pm 1$ \\ 
  wall\_following & 5456 & 24 & 0 & 4 & $99.3 \pm 0.2$ & $99.3 \pm 0.2$ & $\mathbf{99.6 \pm 0.1}$ & $83.4 \pm 0.9$ & $96.3 \pm 0.3$ \\ 
  waveform & 5000 & 21 & 0 & 3 & $\mathbf{79.5 \pm 0.6}$ & $77.9 \pm 0.7$ & $79.2 \pm 0.4$ & $\mathbf{79.5 \pm 0.4}$ & $79 \pm 0.5$ \\ 
  waveform\_noise & 5000 & 40 & 0 & 3 & $79.9 \pm 0.5$ & $78.9 \pm 0.5$ & $79.2 \pm 0.7$ & $79 \pm 0.4$ & $\mathbf{80.3 \pm 0.6}$ \\ 
  wine & 178 & 13 & 0 & 3 & $95 \pm 3$ & $94 \pm 4$ & $\mathbf{97 \pm 2}$ & $96 \pm 2$ & $\mathbf{97 \pm 2}$ \\ 
  wine\_quality-red & 1599 & 11 & 0 & 6 & $\mathbf{47.3 \pm 3}$ & $46.2 \pm 3.4$ & $45.2 \pm 2.9$ & $47.1 \pm 2.6$ & $46.6 \pm 2.7$ \\ 
  wine\_quality-white & 4898 & 11 & 0 & 7 & $\mathbf{43.8 \pm 2.2}$ & $42.8 \pm 1.6$ & $42 \pm 1.2$ & $43.4 \pm 2.2$ & $43.7 \pm 1.9$ \\ 
  yeast & 1484 & 8 & 0 & 10 & $47.7 \pm 1.9$ & $\mathbf{48 \pm 2.5}$ & $47.2 \pm 2.4$ & $46.5 \pm 1.9$ & $46.3 \pm 2.4$ \\ 
  zoo & 101 & 16 & 0 & 7 & $93 \pm 4$ & $\mathbf{94 \pm 3}$ & $93 \pm 2$ & $\mathbf{94 \pm 3}$ & $\mathbf{94 \pm 3}$ \\ 
   \hline
\end{longtable}
\end{landscape}

\end{appendices}



\end{document}